\documentclass[twoside,11pt]{article}

\usepackage[accepted]{melba}

\usepackage{mwe} %

\usepackage{amsmath,amsfonts}

\usepackage{algorithm}
\usepackage{algorithmicx}
\usepackage{booktabs}
\usepackage{graphicx}
\graphicspath{{figures/}}
\DeclareGraphicsExtensions{.pdf,.png,.jpg}
\usepackage[caption=false,font=normalsize,labelfont=sf,textfont=sf]{subfig}
\usepackage{mathtools}%
\usepackage{tikz}
\usetikzlibrary{arrows, automata}

\newcommand{\EE}{\mathbb{E}}
\newcommand{\MSE}{\mathsf{MSE}}
\newcommand{\MNAME}{GLACIAL}

\melbaid{2024:028}  %
\doi{https://doi.org/10.59275/j.melba.2024-c7eg}
\melbaauthors{Nguyen, Ngo, and Sabuncu}  %
\volume{2}
\firstpageno{2223}  %
\melbayear{2024}  %
\datesubmitted{05/2024}  %
\datepublished{11/2024}  %

\ShortHeadings{Granger and Learning-based Causality Analysis for Longitudinal Imaging Studies}{Nguyen et al.}

\title{\MNAME: Granger and Learning-based Causality Analysis for Longitudinal Imaging Studies}

\author{\firstname Minh \surname Nguyen\orcid{0000-0003-4762-1798} \email bn244@cornell.edu \\
	\addr ECE Department, Cornell Tech, New York, NY, USA
	\AND
    \firstname Gia H. \surname Ngo\orcid{0000-0002-2793-2700} \email ghn8@cornell.edu \\
	\addr ECE Department, Cornell Tech, New York, NY, USA
	\AND
	\firstname Mert R. \surname Sabuncu\orcid{0000-0002-7068-719X} \email msabuncu@cornell.edu \\
	\addr ECE Department, Cornell Tech, New York, NY, USA
    \AND for the Alzheimer's Disease Neuroimaging Initiative\thanks{Data used in preparation of this article were obtained from the Alzheimer's Disease Neuroimaging Initiative (ADNI) database (adni.loni.usc.edu). As such, the investigators within the ADNI contributed to the design and implementation of ADNI and/or provided data but did not participate in analysis or writing of this report. A complete listing of ADNI investigators can be found at: http://adni.loni.usc.edu/wp-content/uploads/how\_to\_apply/ADNI\_Acknowledgement\_List.pdf}
}

\begin{document}

\maketitle

\begin{abstract}%
The Granger framework is useful for discovering causal relations in time-varying signals.
However, most Granger causality (GC) methods are developed for densely sampled timeseries data.
A substantially different setting, particularly common in medical imaging, is the longitudinal study design, where \textit{multiple} subjects are followed and sparsely observed over time.
Longitudinal studies commonly track several biomarkers, which are likely governed by nonlinear dynamics that might have subject-specific idiosyncrasies and exhibit both direct and indirect causes.
Furthermore, real-world longitudinal data often suffer from widespread missingness.
GC methods are not well-suited to handle these issues.
In this paper, we propose an approach named \MNAME~(Granger and LeArning-based CausalIty Analysis for Longitudinal studies) to fill this methodological gap by marrying GC with a multi-task neural forecasting model.
\MNAME~treats subjects as independent samples and uses the model's average prediction accuracy on hold-out subjects to probe causal links.
Input dropout and model interpolation are used to efficiently learn nonlinear dynamic relationships between a large number of variables and to handle missing values respectively.
Extensive simulations and experiments on a real longitudinal medical imaging dataset show \MNAME~beating competitive baselines and confirm its utility.
Our code is available at~\url{https://github.com/mnhng/GLACIAL}.
\end{abstract}


\section{Introduction}
Granger causality (GC)~\citep{granger1969investigating} is a versatile and popular framework that exploits ``the arrow of time'' to detect causal relations in timeseries data~\citep{roebroeck2005mapping,zhang2011climate}.
In GC, we test whether past values of one time series predict the future values of another series (i.e., forecasting), which allows us to infer causal relationships.
Despite its popularity, current implementations of GC are only well-suited for densely and uniformly sampled timeseries data from one system at a time.
They are not designed for the longitudinal setup involving multiple systems (e.g., subjects), which are common in medical imaging.
Although one could infer a causal graph for each subject and aggregate the graphs across subjects, this approach is untenable in many longitudinal studies in medicine where each subject only has a few observations, making the inference of each causal graph inaccurate.

Constraint-based methods such as PC or FCI~\citep{spirtes2000causation}, which rely on independent samples and conditional independence tests, are also commonly used for causal discovery.
These methods would use one observation per subject and thus are not designed to detect causal relations reflected in temporal dynamics.
We believe there is a lack of methods for causal discovery in longitudinal studies that consist of multiple subjects with sparse observations.

Longitudinal imaging studies typically track several variables simultaneously.
Thus, applying causal discovery to longitudinal studies can be challenging because of the large number of variables involved and the complex (nonlinear) relationships between variables.
Nonlinear GC methods (e.g.\ those based on non-parametric methods~\citep{su2007consistent,marinazzo2008kernel}) do not scale to large number of variables~\citep{eichler2012causal}.
Similarly, existing GC tests that use neural networks to infer nonlinear dynamics~\citep{tank2021neural,nauta2019causal,khanna2020economy} also face scalability issues.
On the other hand, using linear GC to infer nonlinear relationships can be fast but may produce wrong results~\citep{li2018causal}.

Furthermore, prior GC methods are, to the best of our knowledge, all association-based.
That is, they test for causal relationships via interrogating fit (learned) model weights.
For example, in the linear GC approaches, this is achieved by testing the statistical significance of model coefficients. 
As the detection power of association-based GC~\citep{granger1969investigating,lutkepohl2005new} diminishes with increasing number of variables~\citep{sugihara2012detecting,runge2019detecting}, it may fail to detect the weak coupling between a node and its parents, in particular when there are a lot of variables and limited data~\citep{runge2019inferring,yuan2021data}.
Another challenge of real-world longitudinal studies is missing data.
While there is no consensus about what to do about missing values~\citep{glymour2019review}, several works~\citep{strobl2018fast,tu2019causal} have tried to address this issue for cross-sectional data.
Yet, as far as we know, missingness is under-explored in longitudinal studies, particularly in the context of causal discovery.
Finally, GC, in its original form, does not differentiate between direct and indirect causes~\citep{yuan2021data}.
Although, in theory, infinite history (observations) could shield off indirect causes from being detected as edges in the output causal graph, when the number of observations per subject is small, false positives due to indirect causes is a common practical problem.
\begin{figure}[t]
\centering
\includegraphics[width=.8\linewidth]{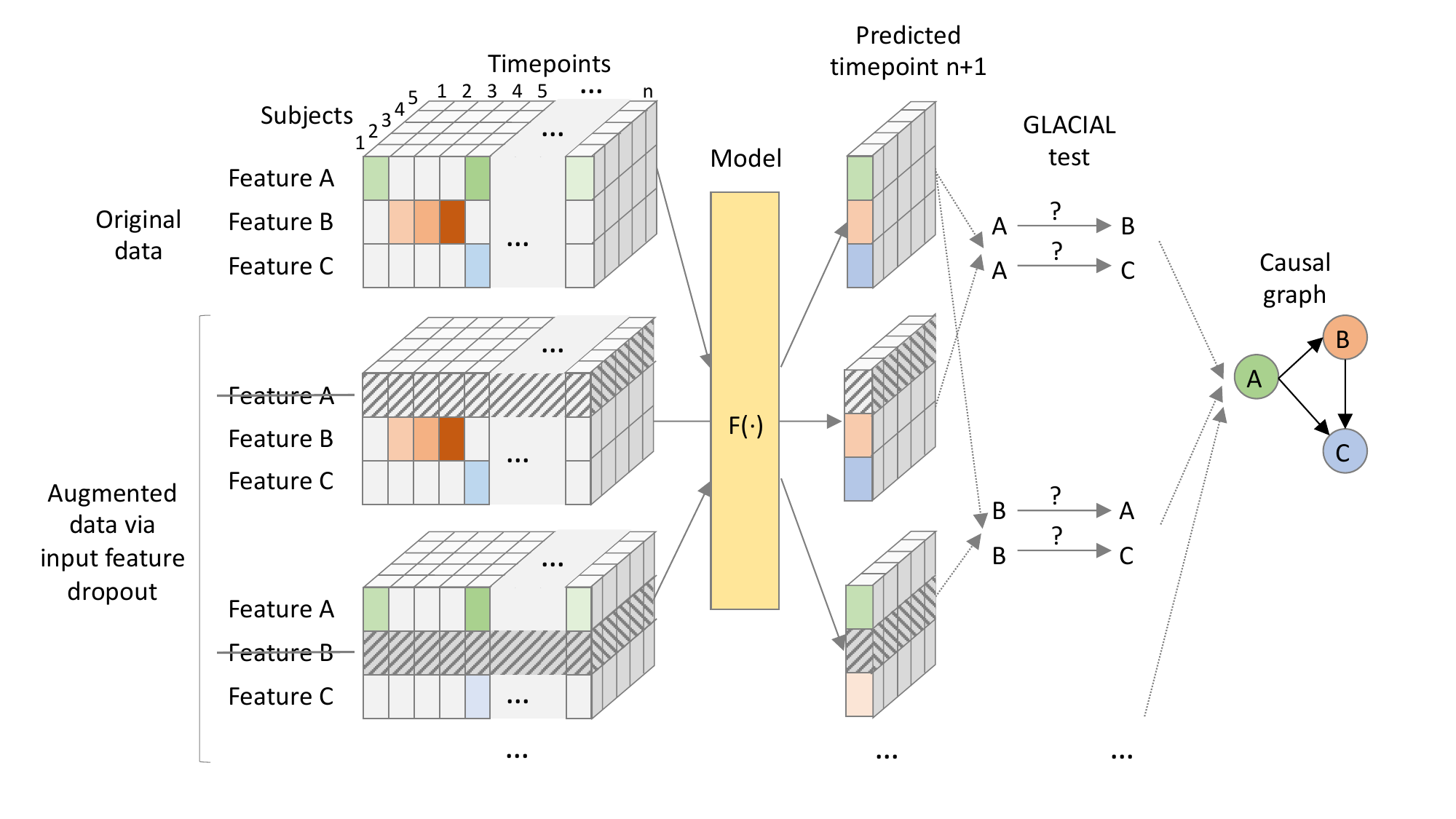}
\caption{\MNAME. Overview of the proposed approach for longitudinal studies.}\label{fig:glacial}
\end{figure}

In this work, we propose \MNAME~(Fig~\ref{fig:glacial}), which stands for a ``Granger and LeArning-based CausalIty Analysis for Longitudinal studies.''
\MNAME~combines GC with a practical machine-learning based approach to test for causal relations among multiple variables in a longitudinal study.
\MNAME~extends GC to longitudinal studies by treating each subject's trajectory as an independent sample, governed by a shared causal mechanism that is reflected in the temporal dynamics.
This treatment is similar to prior works where subjects are assumed to be independent samples in longitudinal data analysis~\citep{hernan2020causal}.
By applying a standard train-test setup with hold-out subjects, \MNAME~can test for effects of causal relations in expectation.
Critically, \MNAME~infers causal relationships based on interrogating predictive accuracy and not a direct analysis of model weights, which is common in existing association-based GC methods.
\MNAME~employs a single multi-task neural forecasting model, trained with input feature drop-out, to learn nonlinear relationships among all variables in time-varying data.
The model also handles missing values using model interpolation.
Thus, although neural networks have been used in the past for causal discovery, \MNAME~efficiently tests for causal relations of a large set of variables in data where timepoints may be sampled irregularly and may contain missing values.
The efficiency and flexibility of \MNAME~make it applicable to real-world multi-modal medical imaging studies with many variables.
Furthermore, \MNAME~includes post-processing heuristics to account for indirect causes and resolve the directionality of detected ambiguous associations.
Extensive experiments on synthetic data and real data from a longitudinal medical imaging study show that \MNAME~can infer relationships accurately even in challenging real-world scenarios with sparse observations, a large number of variables and direct causes, and a large degree of missing data.
Although a specific model was used in our experiments, \MNAME~is model-agnostic.

\section{Related Works}
Most existing causal discovery (CD) methods are not intended for the longitudinal study design, where multiple subjects are sparsely observed at different timepoints.
CD methods designed for timeseries data or independent samples are often used in the longitudinal setting despite potential poor performance.

\textbf{Causal Discovery:}
CD methods intended for cross-sectional studies are ill-suited for longitudinal studies.
They often fall under: constraint-based search (e.g. FCI~\citep{spirtes2000causation}), score-based search (e.g. Greedy Equivalence Search (GES)~\citep{chickering2002optimal}), functional causal models (FCMs)~\citep{shimizu2006linear,hoyer2008nonlinear,zhang2009additive,zhang2006extensions,zhang2009identifiability}, or continuous optimization~\citep{zheng2018dags}.
Search methods can scale well if causal relations are linear~\citep{kalisch2007estimating,ramsey2017million} although their output may not be informative enough (e.g.\ containing bidirectional edges).
In contrast, by making strong assumptions about the functional form of the causal process, FCM can better identify the causal direction~\citep{hyvarinen1999nonlinear,zhang2015estimation}, although FCM methods usually do not scale well~\citep{glymour2019review}.
Besides, if the assumed FCM is too restrictive to be able to approximate the true data generating process, the results may be misleading.

There are also various CD methods for timeseries~\citep{chu2008search,runge2019detecting,runge2020discovering,entner2010causal,malinsky2018causal,malinsky2019learning,hyvarinen2010estimation,peters2013timino,pamfil2020dynotears}.
These methods take in consecutive blocks of observations and output a Full Time Graph~\citep{peters2017elements}, which contain not only the variables in the system but also their temporally-lagged versions.
Although methods for timeseries may be better than cross-sectional ones, they are still not ideal for longitudinal data where sparse observations with potentially missing values come from more than one subject.

\textbf{Granger Causality:}
GC~\citep{granger1969investigating,granger1980testing} checks for dependence between variables' timeseries, after accounting for other available information.
Temporal dependence is thus linked to causation by the ``Common Cause Principle'': two dependent variables are causally related (one causes the other, or both share a common cause)~\citep{peters2017elements}.
Checking pairwise dependence in GC can be efficient, but often yields false positives because other variables in the system are not accounted for.
In contrast, multivariate GC can account for common causes and therefore is more accurate but also more computationally demanding~\citep{eichler2007granger,eichler2012causal}.
In practice, multivariate GC may be infeasible for a large set of variables and more efficient approaches~\citep{basu2015network,huang2015fast} were developed to deal with this challenge.
Recently, more general GC tests based on neural networks~\citep{tank2021neural,nauta2019causal,khanna2020economy} have been proposed which outperform vector auto regressive (VAR) linear GC~\citep{glymour2019review}.
Scaling these neural-network based GC methods to handle a large number of variables is still a concern.

\textbf{Missing data:}
For cross-sectional studies, missing values can be imputed, which may result in data contradicting the causal processes if imputation is done naively.
Alternatively, observations with missing values can be removed (list-wise deletion), which can lead to the omission of vast amounts of valuable datapoints.
Test-wise Deletion PC (TDPC)~\citep{strobl2018fast} is more data-efficient than list-wise deletion but may produce spurious edges when missingness is not completely at random~\citep{tu2019causal}.
Missing-Value PC (MVPC)~\citep{tu2019causal} corrects TDPC's output to account for different missingness scenarios.
Although, tackling missingness in data through imputation has been studied extensively~\citep{ma2019eddi,ma2020vaem,ma2021identifiable,morales2022simultaneous} in the context of cross-sectional studies, to the best of our knowledge, no existing method addresses missingness in longitudinal studies for the CD task.

\section{Method}\label{sec:method}
Both cross-sectional CD methods (multiple subject, single timepoint data) and timeseries CD methods (single subject, multiple timepoints data) are ill-suited for longitudinal studies (multiple subjects, multiple timepoints data).
Besides, prior methods often assume timeseries are infinitely long (i.e.\ unlimited history), regularly sampled, and without missing values.
Thus, they may not work for real-world datasets when observation history per subject is limited, irregular, and riddled with missing values.
The next few sections show (1) how \MNAME~handles longitudinal data, (2) how \MNAME~deals with irregularly sampled timepoints containing missing values, and (3) \MNAME's post-processing strategies to account for limited history of observed timeseries.

Causal discovery is impossible without assumptions.
\MNAME~assumes causal faithfulness, no hidden confounder, acyclicity (DAG, hence no feedback effect) and no instantaneous effects (the first three assumptions are standard in CD literature, c.f.~\citep{pearl2009causality}).
\MNAME~does not assume stationarity unlike linear GC.

\subsection{Longitudinal Study Set-up}
In a longitudinal imaging study, there are \textit{multiple} subjects who are sparsely scanned during a limited number of visits.
Let $\mathbf{X}_t$ and $\mathbf{Y}_t$ be time-varying variables (e.g., two image-derived biomarkers, or an image-derived biomarker and a clinical score) indexed with non-negative integer $t \in \{0, \ldots T{-}1\} = [T]$.
We use super-script notation to indicate history: $\mathbf{X}^t=\{\mathbf{X}_{0}, \ldots, \mathbf{X}_{t-1}\}$.
$\mathbf{\Omega}^t = \mathbf{X}^t \cup \mathbf{Y}^t \cup \dots$ is the union of histories of all variables.
The data from subject $i$ with $T_i$ observations ($T_i\leq T$) is $\mathbf{\Omega}^{T_i}$.
The whole longitudinal dataset is $\{\mathbf{\Omega}^{T_i};\;\; i\in 1,\ldots, N\}$.
The number of observations, $T_i$, is usually less than 10 (sparse) and can be as low as 1 or 2.
The number of subjects, $N$, is often less than 10,000.
The $\mathbf{\Omega}^{T_i}$ matrices may contain missing values.

\subsection{Granger Causality Formulation}\label{ssec:gc_formulation}
A popular GC test is based on comparing the mean squared error (MSE) achieved by two predictors~\citep{granger1980testing}.
In the GC MSE formulation, we conclude that ``$Y$ causes $X$'' if:
\begin{align}
\delta_t(X|Y) &= \MSE(\mathbf{X}_t, \EE[\mathbf{X}_t | \mathbf{\Omega}^{t} \setminus \mathbf{Y}^{t}]) \nonumber\\
              &- \MSE(\mathbf{X}_t, \EE[\mathbf{X}_t | \mathbf{\Omega}^{t}])  > 0, \forall t \in [T] \label{eq:mse}
\end{align}
where $\EE$ denotes (conditional) expectations.
Eq~\ref{eq:mse} simply calculates the MSE difference between two optimal (in an MSE sense) predictors of $X$ (see Appendix~\ref{app:gc_mse} and~\citep{granger1980testing}).
The first predictor (i.e. $\EE[\mathbf{X}_t | \mathbf{\Omega}^{t} \setminus \mathbf{Y}^{t}]$) is not given information about $Y$.
The second predictor (i.e. $\EE[\mathbf{X}_t | \mathbf{\Omega}^{t}]$) is given all past information, including about $Y$.
Since $\delta_t(X|Y) > 0, \forall t$, Eq~\ref{eq:mse} can be adapted for longitudinal data as:
\begin{align}
    \Delta\MSE(X|Y) = \EE_i\Big[\frac{1}{T_i} \sum_{t=0}^{T_i-1}\delta_t(X|Y)\Big] > 0 \label{eq:pop_mse}
\end{align}

Relying on the assumption that statistical dependence implies a causal link~\citep{reichenbach1956direction}, when past values of $Y$ predict future values of $X$ (dependence): (1) $X$ causes $Y$ OR (2) $Y$ causes $X$ OR (3) $X$ and $Y$ have a common cause.
With sufficiently high sampling rate, causes are observed to occur before effects in time, thus ruling out (1).
The no hidden confounder assumption rules out (3).
Hence, a positive test implies that ``$Y$ causes $X$''.
The next section details how this test can be done in practice when the optimal predictors are not given.
We are particularly interested in the setting with multiple observed independent subject trajectories.

\subsection{Choice of Predictor}\label{ssec:imp_details}
We can approximate the MSE-optimal predictors with neural networks $F$ and $G$.
\begin{align}
    \delta_t(X|Y;F,G) &= \MSE(\mathbf{X}_t, F(\mathbf{X}_t; \mathbf{\Omega}^{t} \setminus \mathbf{Y}^{t})) \nonumber\\
    &- \MSE(\mathbf{X}_t, G(\mathbf{X}_t; \mathbf{\Omega}^{t})) \label{eq:mse_nn}\\
    \Delta\MSE(X|Y;F,G) &= \EE_i\Big[\frac{1}{T_i} \sum_{t=1}^{T_i}\delta_t(X|Y;F,G)\Big] \label{eq:pop_mse_nn}
\end{align}

To calculate $\delta_t(X|Y;F,G)$, we first have to train the forecasting neural networks.
Once trained, the neural networks can be used to calculate $\Delta\MSE(X|Y;F,G)$ on hold-out test subjects.
Thus, the predictors' performance depends on the training data, optimization, network initialization, and other implementation details.
Even with the best optimizer and initialization procedure, a bad training-test split could, for instance, result in a sub-optimal model and consequently false causal link estimates.
For more robust causal discovery, in \MNAME, we repeat the estimation of $\Delta\MSE(X|Y;F,G)$ multiple times using different random splits of data and test that $\Delta\MSE$ is positive on average using a statistical test.

We use a \textit{single} forecasting recurrent neural network (RNN)~\citep{graves2009novel} in place of all predictors.
The RNN is trained to predict the next step values of all the variables, $\mathbf{\Omega}_t$, given all available past values, $\mathbf{\Omega}^{t}$.
We adopt the RNN model from~\citep{nguyen2020predicting} since it implements model interpolation to handle missing values.
In particular, if there are missing values at time $t$, they can be replaced by the RNN prediction, $\mathbf{\widehat{\Omega}}_t$ (model interpolation).
This timepoint with interpolated values is then concatenated with $\mathbf{\Omega}^{t}$ to form $\mathbf{\Omega}^{t+1}$ which is subsequently used to predict values at time $t{+}1$.
In this way, the RNN is not aware whether the features are observed/measured or imputed.
Missing values are ignored when calculating training loss and estimating $\delta_t(X|Y;F,G)$ on hold-out subjects (Eq~\ref{eq:mse_nn}).
Training follows~\citep{nguyen2020predicting} so that even data containing missing values can be used for RNN training.
Note that the choice of the neural network is not very critical.
Given that the network can fit the data well, any neural network model that forecasts future values from past values and implements model interpolation should work in \MNAME.

\textbf{Input Feature Dropout:}
Training separate neural networks to compute $\Delta\MSE(X|Y; F, G)$ for each variable pair would make this approach infeasible for medical imaging studies with numerous variables.
This is because the number of networks required would be proportional to the number of variables squared.
Instead, we propose to train a single multi-task (i.e., multi-output) RNN, $F(\cdot;\theta)$, to approximate $\EE[\mathbf{X}_t | \mathbf{\Omega}^{t} \setminus \mathbf{Y}^{t}]$ and $\EE[\mathbf{X}_t | \mathbf{\Omega}^{t}]$, for all predicted variables $\mathbf{X}_t$.
A similar technique has been shown to allow a single model to learn multiple predictive functions~\citep{nguyen2024knockout}.
The RNN acts as the former when $Y$ is masked out of the input vector and acts as the latter when the input is complete.
To obtain a model that can produce accurate predictions under these scenarios, during training, we augment each mini-batch by dropping out subject variables from the input features.

\textbf{Implementation Details:}
The same settings of \MNAME~are used in all experiments.
We used repeated 5-fold cross-validation to split a dataset into training, validation, and test sets with a 3:1:1 ratio.
The RNN is trained to minimize next-step prediction error using Adam~\citep{kingma2015adam}, L2 loss, and a learning rate of 3E-4.
The RNN has one hidden layer of size 256.
Training was done on an NVIDIA TITAN Xp GPU.
The validation set is used for early stopping.
Cross-validation is repeated 4 times, resulting in 20 different splits of data.
We find 4 repetitions to strike a good balance between robustness and speed.
Running more repetitions might slightly improve the results when missingness is severe but at a higher computational cost (see Appendix~\ref{app:more_reps}).
We perform a t-test on the $\Delta\MSE$ statistic and use the significance level threshold of 0.05.

\subsection{Post-Processing}\label{ssec:postproc}
GC assumes history of the timeseries is infinite.
When observations are finite as in real-world longitudinal studies, GC may draw wrong conclusions.
E.g., consider following deterministic system:
\begin{align*}
    \textbf{Y}_t &= a \textbf{Y}_{t-1} + b \textbf{Y}_{t-2} \\
    \textbf{X}_t &= c \textbf{Y}_{t-2}.
\end{align*}
In this system, $Y$ causes $X$ since manipulating $Y$ will change the value of $X$.
By the same logic, $X$ is not the cause of $Y$ because manipulating $X$ will not change $Y$.

When history is infinite, GC works as expected
\begin{align*}
    &\EE[\mathbf{Y}_t | \mathbf{X}^{t}, \mathbf{Y}^{t}] = \EE[\mathbf{Y}_t | \mathbf{Y}^{t}] = \mathbf{Y}_t \\
    &\MSE(\mathbf{Y}_t, \EE[\mathbf{Y}_t | \mathbf{Y}^{t}]) = \MSE(\mathbf{Y}_t, \EE[\mathbf{Y}_t | \mathbf{X}^{t},\mathbf{Y}^{t}]) = 0 \\
    &\Rightarrow X \;\text{does not cause}\; Y \quad\text{(correct)}
\end{align*}
However, when only 1 past timepoint is given (finite history), GC draws a wrong inference.
\begin{align*}
&\MSE(\mathbf{Y}_t, \EE[\mathbf{Y}_t | \mathbf{Y}_{t-1}])
  > \MSE(\mathbf{Y}_t, \EE[\mathbf{Y}_t | c\mathbf{Y}_{t-2},\mathbf{Y}_{t-1}])
 = \MSE(\mathbf{Y}_t, \EE[\mathbf{Y}_t | \mathbf{X}_{t-1},\mathbf{Y}_{t-1}]) \\
&\Rightarrow X \;\text{causes}\; Y \quad\text{(incorrect)}
\end{align*}
Thus, GC may detect edges in both direction ($X \rightarrow Y$ and $Y \rightarrow X$) for a pair of variables when limited history is given.
It can be shown in a similar fashion that if $X$ causes $Y$ and $Y$ causes $Z$ ($X$ is the indirect cause of $Z$), $Y$ will not be able to shield $Z$ from $X$ if only limited history is given.
Thus, GC will also detect edges for indirect causes in both direction ($X \rightarrow Z$ and $Z \rightarrow X$).

\MNAME~includes two additional post-processing steps to remove these false positives.
Let $S(X|Y)$ be the statistic (e.g.\ the t-test) that tests for the positivity of $\Delta\MSE$ from several train/test splits.
Thus $S(X|Y)$ can be viewed as a test for whether $Y$ causes $X$.

\textbf{1. Orient bidirectional edge:}
If $S(X|Y) < S(Y|X)$ remove $Y{\rightarrow}X$, else remove $X{\rightarrow}Y$.
This step is similar to prior work such as~\citep{hoyer2008nonlinear,zhang2009additive,janzing2012information,kocaoglu2017entropic} which leverages causal asymmetry to determine the causal direction (the direction with the bigger effect is regarded as the causal direction).
T-statistic has been shown to be informative for causal discovery~\citep{weichwald2020causal}.
Appendix~\ref{app:first_heuristic} presents a mathematical justification for this heuristic.

\textbf{2. Remove indirect edge:}
Remove edge $X{\rightarrow}Y$ if there exists an alternative path $(X{:=}U_0,U_1,\dots,Y{:=}U_k)$ from $X$ to $Y$ or a path $(Y{:=}U_0,U_1,\dots,X{:=}U_k)$ from $Y$ to $X$ if \[ S(Y|X) < \min(\{S(U_{j+1}|U_j);\;\; j\in0,\dots,k{-}1\}). \]
Intuitively, if there is an alternative path on which the effect of the weakest edge is greater than the effect of $X{\rightarrow}Y$ then $X$ is likely an indirect cause of $Y$.
A complete description of \MNAME~is shown in Algorithm~\ref{algo:algo}.

\begin{algorithm}[t]
\caption{\MNAME}\label{algo:algo}
\begin{algorithmic}
\State \textbf{In}: Data splits $(D^{train}_1,D^{test}_1),\dots,(D^{train}_n,D^{test}_n)$
\State \textbf{Out}: Causal graph $G$
\State \textsc{Step 1: Association check using the GC MSE test}
\State \hspace{0.0cm} \textbf{For each} data split $D_i$
\State \hspace{0.3cm}     Fit RNN model $F_i$ using $D^{train}_i$\;
\State \hspace{0.3cm}     \textbf{For each} variable pair $(u, v)$
\State \hspace{0.6cm}         Calculate $\Delta$MSE[u, v, i] using $F_i$ and $D^{test}_i$;
\State \hspace{0.0cm} \textbf{For each} variable pair $(u, v)$
\State \hspace{0.3cm}     t-statistic, p-value = t-test($\Delta$MSE[u, v, $\ast$]);  $\quad$ (t-test across data splits)
\State \hspace{0.3cm}     \textbf{If} p-value $<$ threshold
\State \hspace{0.6cm}         Add $u{\rightarrow}v$ to $G$; $\quad$ S[u, v] = t-statistic;
\State \textsc{Step 2: Orient bidirectional edges}
\State \hspace{0.0cm} \textbf{For each} bidirectional pair $u{\rightarrow}v$ and $v{\rightarrow}u$ in $G$
\State \hspace{0.3cm}     \textbf{If} {$S[u, v] < S[v, u]$}
\State \hspace{0.6cm}         Remove $u{\rightarrow}v$ from $G$; ($v{\rightarrow}u$ has stronger effect)
\State \hspace{0.3cm}     \textbf{Else}
\State \hspace{0.6cm}         Remove $v{\rightarrow}u$ from $G$; ($u{\rightarrow}v$ has stronger effect)
\State \textsc{Step 3: Prune indirect causes}
\State \hspace{0.0cm} \textbf{For each} $u{\rightarrow}v$ in $G$
\State \hspace{0.3cm}     \textbf{For each} path $p=(u{=}w_0, w_1,\dots,w_k{=}v)$
\State \hspace{0.6cm}         \textbf{If} {$S[u, v] < S[w_j, w_{j+1}]\;\;\forall j\in\{0,\dots,k{-}1\}$}
\State \hspace{0.9cm}             Remove $u{\rightarrow}v$; $\quad$ break;
\State \hspace{0.3cm}     \textbf{For each} path $p=(v{=}w_0, w_1,\dots,w_k{=}u)$
\State \hspace{0.6cm}         \textbf{If} {$S[u, v] < S[w_j, w_{j+1}]\;\;\forall j\in\{0,\dots,k{-}1\}$}
\State \hspace{0.9cm}               Remove $u{\rightarrow}v$; $\quad$ break;
\end{algorithmic}
\end{algorithm}

\subsection{Runtime Complexity}
Since \MNAME~uses a single multi-task RNN to check relationships between all variable pairs, the number of RNNs trained by \MNAME~is independent of the number of variables and is equal to the number of data splits (see Algorithm~\ref{algo:algo} STEP 1).
For example, with 4 repetitions of 5-fold cross-validation, \MNAME~needs to train 20 different RNNs.
This number is the same whether there are 10 or 100 variables.
Obviously, having more variables will lead to longer execution time per batch but much of the computation is parallelizable (as long as the batch fits into GPU memory).
Therefore, the runtime complexity is mostly dominated by the number of RNNs that must be trained.

\section{Experiments}
\subsection{Experimental Set-up}
In addition to the problems listed in Section~\ref{sec:method}, CD methods often struggle when (1) relationships are non-linear, (2) the number of variables is large, or (3) a node has many parents.
The subsequent experiments are designed to show \MNAME's efficacy and to show that \MNAME~is less affected by these problems.
First, the simulations include both non-linear trajectories and linear random-walk trajectories.
\MNAME~is also applied on real multivariate medical imaging data which most likely include non-linear trajectories.
Second, there is a simulation with a moderate-size graph consisting of 39 nodes to demonstrate scalability.
Third, the simulation with the 39-node graph includes one node (i.e. 22) with 18 direct causes.

\subsubsection{Baselines}
We benchmark \MNAME~against both CD methods for cross-sectional data and CD methods for timeseries.
Only representative and competitive baselines are shown (see Appendix~\ref{app:more_comparisons} for the remaining baselines).

\textbf{CD Methods for Cross-Sectional Data:}
We compare against PC, FCI~\citep{spirtes2000causation}, GFCI~\citep{ogarrio2016hybrid}, and \textit{Sort-N-Regress}~\citep{reisach2021beware} (SnR).
GFCI combines GES and FCI into a single algorithm.
SnR is a simple baseline to ensure that benchmarked approaches go beyond exploiting differences in variables' marginal variance~\citep{reisach2021beware}.
As these approaches assume independent observations, only first timepoints (observations) of subjects are used.
In most longitudinal studies, subjects are guaranteed to have first timepoints (but not other timepoints).
Hence, using the first timepoints will result in the most number of independent timepoints with the least amount of missing data in real-world datasets.
Besides, using all timepoints led to worse performance in our preliminary experiments using simulated data.
Similar to~\citep{shen2020challenges}, GFCI is run multiple times (i.e.\ 20) using different bootstraps of subjects' first timepoints, resulting in multiple graphs.
Only edges appearing in more than half of the resultant graphs are kept in final graph.
Using a higher threshold (80\%) led to worse result (see Appendix~\ref{app:higher_threshold}).

\textbf{CD Methods for Timeseries Data:}
We also adopt SVAR-GFCI~\citep{malinsky2019learning}, PCMCI+~\citep{runge2020discovering}, DYNO-TEARS~\citep{pamfil2020dynotears}, and several GC-based approaches as baselines.
GC-based approaches include linear GC and more recent neural GC tests: cMLP and cLSTM from~\citep{tank2021neural}, TCDF from~\citep{nauta2019causal}, SRU and eSRU from~\citep{khanna2020economy}.
For linear GC, F-statistic was used to test for presence of edges using the same threshold as in \MNAME.
For longitudinal data, one could either (1) estimate one causal graph for each subject and aggregate the graphs or (2) estimate just one graph using concatenated data from all the subjects.
Since (1) often fails when the number of timepoints per subject is sparse, (2) was used instead.
Causal discovery using concatenated subjects' data has been investigated in~\citep{di2019interregional,qing2021causal}.
Besides, linear GC could output false positives when timeseries are non-stationary~\citep{he2001spurious}.
One could make the timeseries stationary by calculating the difference~\citep{evgenidis2017towards} or the log difference between timepoints~\citep{stock2012disentangling}.
However, using differences led to worse results so we report the results using the original timeseries instead.

The input to SVAR-GFCI and PCMCI+ are also the concatenated timeseries from all the subjects.
For DYNOTEARS which can accept timeseries from multiple subjects, the timeseries are not concatenated.
The hyper-parameters of SVAR-GFCI, PCMCI+, DYNOTEARS and neural GC tests are selected based on the suggestions in their original publications.

\textbf{Missing data:}
For data with missingness, TDPC~\citep{strobl2018fast} and MVPC~\citep{tu2019causal} are used instead of GFCI.
For a dataset, each algorithm is run 20 times and the results are aggregated using the same 50\% threshold.
As far as we know, there is no prior work on applying causal discovery methods to timeseries data with missing values.
Therefore, we used linear interpolation to fill out missing values in the data before applying these methods (linear/neural GC, SVAR-GFCI, PCMCI+, and DYNOTEARS).
It may not be feasible to apply more complex interpolation methods since the number of timepoints of a subject can be as low as 2 (after discounting missing values).

\subsubsection{Simulated Data}\label{ssec:sim_data}
The sample size in the simulations was set to 2000 subjects, roughly the size of the ADNI dataset.
Only six timepoints are extracted from each subject's timeseries to simulate sparse observations (see Appendix~\ref{app:more_timepoints} for results with 24 timepoints).
We consider two scenarios.
First, the temporal dynamics are parameterized via the sigmoid function, which is a widely used model for the trajectories of biomarkers, e.g., in Alzheimer's disease~\citep{jack2013tracking}.
In the second scenario, we implement random-walk series.
We experimented with 3 different structural causal models (SCMs), one linear SCM and two non-linear SCMs.
See Appendix~\ref{app:data_gen} for further details.
Results for the non-linear SCMs are shown in Appendix~\ref{app:non_linear_scm}.
As causal structure of simulated data may leak through variables' marginal variance, the data are standardized to zero-mean and unit-variance to prevent CD algorithms from gaming the simulated data~\citep{reisach2021beware}.

\begin{figure}[!h]
\centering
\includegraphics[width=.95\linewidth]{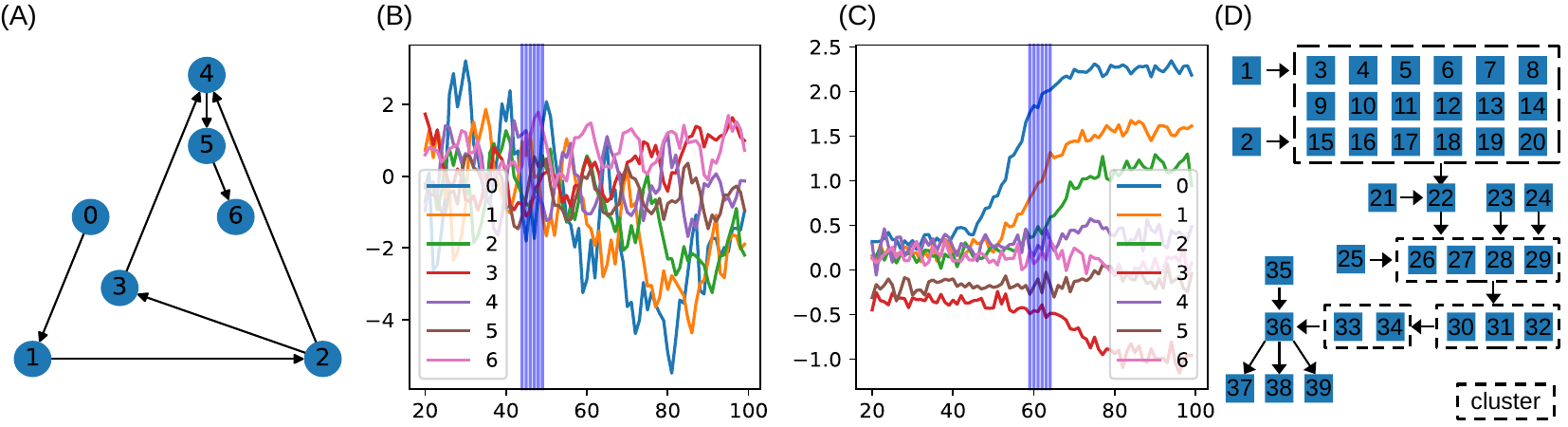}
\caption{Simulation.
(A) 7-node graph having all basic structures (chain, fork, collider).
(B) Subject with random-walk trajectories and linear SCM (data before standardizing to zero mean and unit variance). Only timepoints under vertical lines are observed.
(C) Subject with sigmoid trajectories and linear SCM.
(D) More realistic 39-node graph resembling the RTK/RAS signaling pathway. Nodes in the same cluster have the same causal relations.
}\label{fig:graphs}
\end{figure}

Fig~\ref{fig:graphs} shows the causal graphs used for generating the synthetic data.
The first graph (7 nodes) contains all the basic structures, namely chain, fork, and collider.
The second graph (39 nodes) is used to demonstrate \MNAME's scalability.
This graph is inspired by the RTK/RAS signaling pathway in oncology and is taken from~\citep{sanchez2018oncogenic}.
The second graph is a realistic target that a causal discovery algorithm should be able to find from observational data.
Since the shape of the evolution of signaling proteins is not known, we use Gaussian random-walk as the sample path function.
To simulate missingness (completely at random; MCAR), the values for each timeseries of a subject are independently dropped at fixed rate $p\in\{0.1,0.3,0.5\}$.
Since real data missingness may be more adverse than MCAR, results on simulated missing data are optimistic estimates of performance.
The missingness rate is chosen to match the rate in real data.
Since values from different timeseries are dropped independently, the resulted data could contain subjects with all timepoints having at least one missing values. 

\subsubsection{Real-world Data from an Alzheimer's Disease Study}\label{ssec:adni_data}
We use ADNI~\citep{jack2008alzheimer}, a longitudinal study of Alzheimer's disease (AD) that consists of 1789 subjects and includes multi-modalities of medical images.
Each subject in ADNI has about 7 timepoints on average.
The ADNI study tracks multiple AD biomarkers such as region-of-interest (ROI) volumes (e.g.~hippocampal) derived from structural MRI scans, cognitive tests (e.g. ADAS13), proteins (e.g.~amyloid beta) derived from cerebral spinal fluid samples, and molecular imaging that captures the brain's metabolism (e.g.~FDG PET).
The missingness rates vary for different biomarkers, ranging from $30\%$ (ADAS13) to around $80\%$ (FDG PET).
The variables are shown in Fig~\ref{fig:adni_12var} and described in Appendix~\ref{app:adni}.
For ease of interpretability, we only experiment with summary statistics (volumetric measurements) of medical images.
However, \MNAME~can be easily extended to find causal relations between more fine-grain variables in medical images.

\subsubsection{Metrics}
F1-score, which is the harmonic mean of precision and recall, is used to quantify different approaches' performance.
Note that we assume that there is a ground-truth (directed) graph that describes causal relations.
Each method will also return a list of directed edges between variables.
Precision is the ratio of correctly identified edges over all predicted edges, while recall is the ratio of correct edges over all ground-truth edges.
A predicted edge is considered incorrect if the edge does not exist in the ground-truth graph or the predicted direction contradicts the ground-truth direction.
Thus, a predicted bidirectional edge would be incorrect if the ground-truth edge has only one direction.

\section{Results}
\subsection{Simulated Data}\label{ssec:sim_result}
\begin{figure}[!h]
\centering
\includegraphics[width=.95\linewidth]{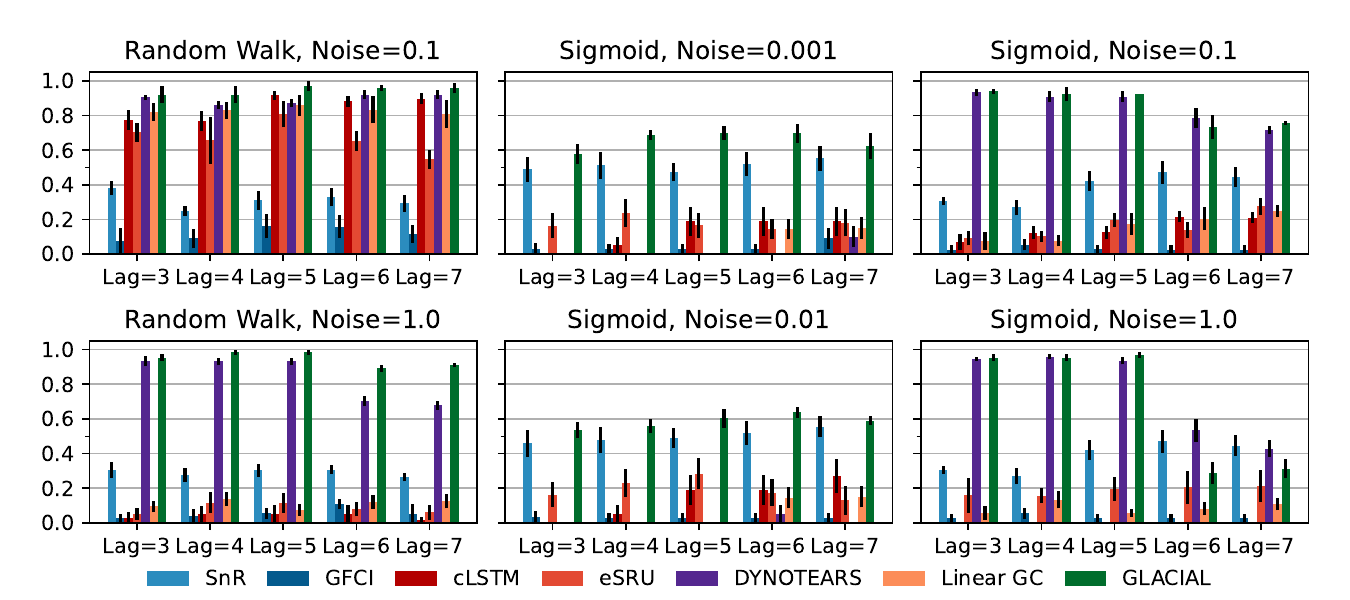}
\caption{Average F1-scores at different settings of sample path, lag-time and measurement noise (7-node graph).
\MNAME~outperforms baselines in most settings (see Appendix~\ref{app:more_comparisons} for more comparisons).
}\label{fig:main_7n1}
\end{figure}

\textbf{7-node graph:}
For random-walk data, \MNAME~outperforms the baselines for various lag-times and measurement noise levels (Fig~\ref{fig:main_7n1}, 1st column).
Similarly, \MNAME~also outperforms the baselines, for the sigmoid data (2nd and 3rd column).
\MNAME's performance dips (3rd column) when input history (5 years) is shorter than the lag-time (6 or 7 years).
This dip is more pronounced when measurement noise is high (3rd column, bottom).

DYNOTEARS fails to detect causal relations in systems with almost deterministic dynamics (2nd column) even though it is the best baseline.
System with deterministic dynamics is also challenging for linear GC~\citep{peters2017elements} although it is slightly better than DYNOTEARS (F1-score $<$ 0.2).
Interestingly, \MNAME~still works in these systems (F1-score = 0.6).
Only \MNAME~manages to consistently beat the strong SnR (\textit{Sort-N-Regress}) baseline.

\begin{figure}[!h]
\centering
\includegraphics[width=.95\linewidth]{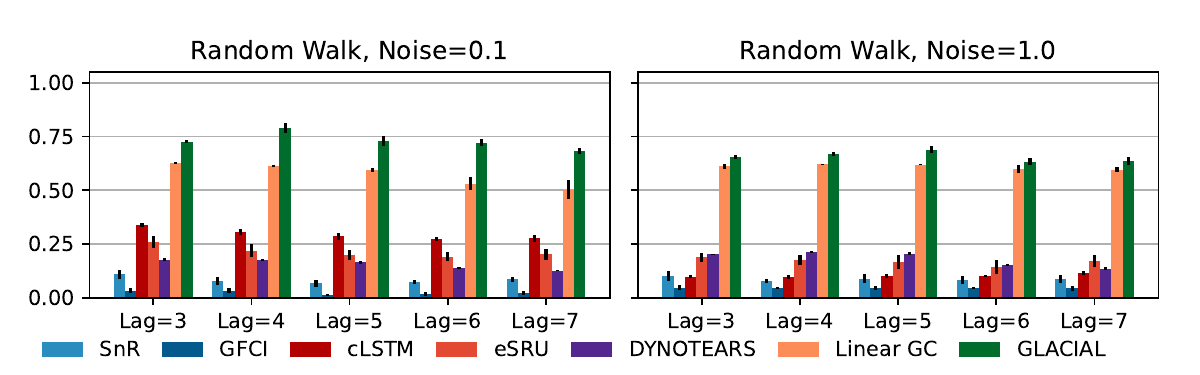}
\caption{Average F1-scores at different settings of lag-time and measurement noise (39-node graph, Gaussian random-walk).
\MNAME~outperforms baselines in most settings (see Appendix~\ref{app:more_comparisons} for more comparisons).
}\label{fig:main_bro_rtk}
\end{figure}

\textbf{39-node graph:}
Although DYNOTEARS is the best baseline for the 7-node graph, its performance on the big graph is worse than linear Granger (Fig.~\ref{fig:main_bro_rtk}).
\MNAME~consistently outperforms all baselines on this large graph when the sample path is Gaussian random-walk.
\MNAME~performs quite well despite the presence of a cluster of direct causes whose contribution to node ``22'' may be too small to detect.

\begin{figure}[!h]
\centering
\includegraphics[width=.95\linewidth]{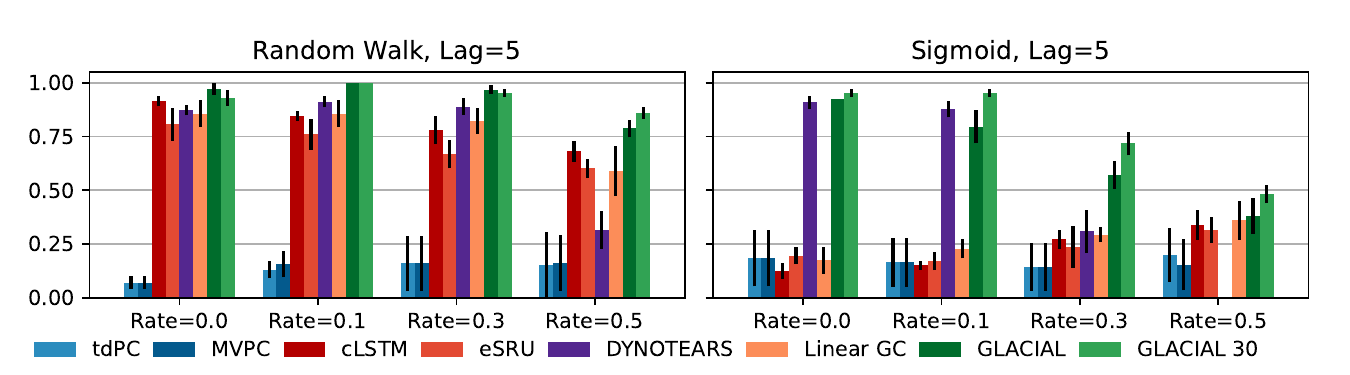}
\caption{Average F1-scores at various levels of missing at random. Lag-time=5. Noise level = 0.1.
\MNAME~usually outperforms baselines.
Running \MNAME~for more repetitions (i.e. 30 instead of 4, denoted as \MNAME~30; see Section~\ref{ssec:imp_details}) can improve performance when dealing with missing data.
}\label{fig:missing}
\end{figure}

\textbf{Missing data:}
Fig~\ref{fig:missing} shows F1-scores at different degrees of missingness.
\MNAME~outperforms TDPC and MVPC, CD approaches tailored for missing data, by better exploiting the temporal dynamics within subjects' timeseries.
\MNAME~also outperforms CD methods for timeseries such as cLSTM and DYNOTEARS.
Although being the best baseline, DYNOTEARS often fails when the missingness level is high (>0.3).
When half of the values are missing (=0.5), \MNAME~can still infer some causal relations.
As an aside, \MNAME's performance on missing data can be improved with more repetitions (see Appendix~\ref{app:more_reps}).

\subsection{\MNAME's Post-processing Ablation}\label{ssec:postproc_abl}
\begin{figure}[!h]
    \centering
    \includegraphics[width=.95\linewidth]{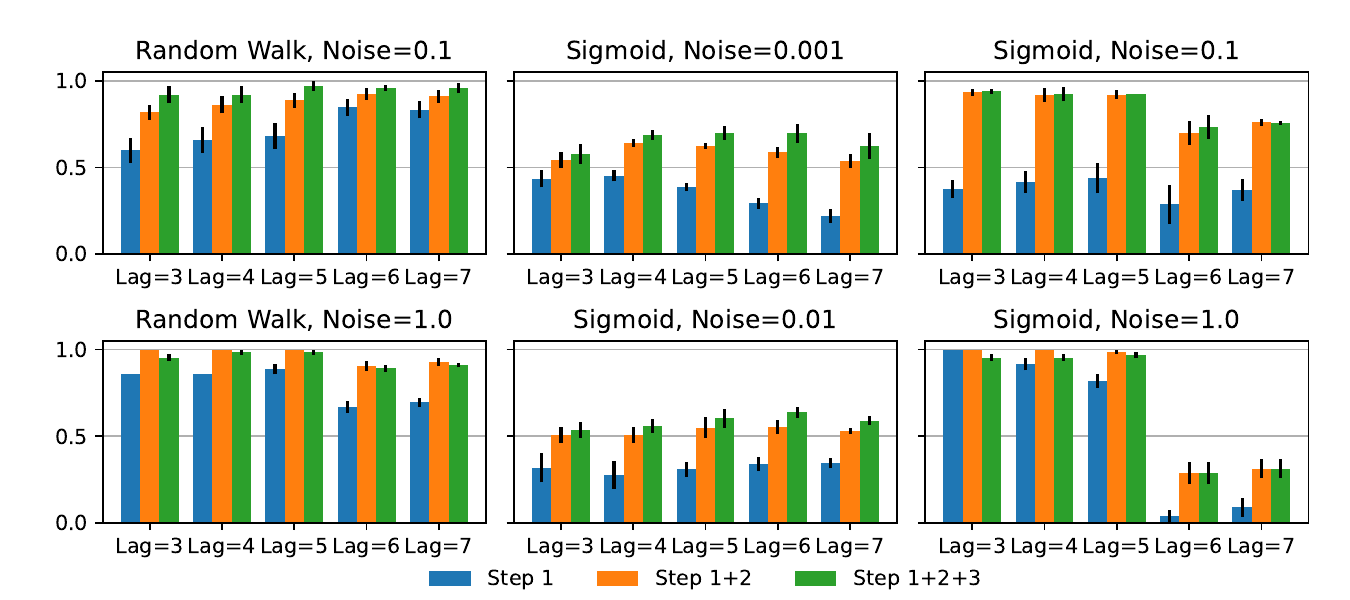}
    \caption{Contribution from \MNAME's heuristics to F1-scores. 7-node graph simulation.}\label{fig:ablate_stages}
\end{figure}

\MNAME's first step tests for edges in the causal graph by comparing the difference in MSE on hold-out subjects.
However, when test subjects are only sparsely observed for a limited number of times, this step may find spurious edges (edges from effect to cause or edges between indirect cause, e.g.\ a grand-parent, and effect).
To address this problem, \MNAME~has two additional heuristics: one (Step 2) to remove edges from effect to cause and another (Step 3) to prune edges between indirect cause and effect.
Fig~\ref{fig:ablate_stages} shows the contribution of these two post-processing heuristics to F1-scores at various lag-times and noise levels (7-node graph simulation).
The first heuristic (Step 2) consistently leads to better results.
While the second heuristic (Step 3) is beneficial most of the time, it can sometime result in performance degradation.
Thus, when applying \MNAME~to real data, it is recommended to compare the outputs with and without the second heuristic to decide which output is more plausible.

\subsection{\MNAME's Hyper-parameter Sensitivity Ablation}\label{ssec:hyperparam_sensitivity}
Since \MNAME~uses neural network for inference, one may think that its results are sensitive to the choice of hyper-parameters.
We analyzed \MNAME's performance as the hyper-parameters vary.
Table~\ref{tbl:arch} shows the performance of \MNAME~while varying (1) the number of hidden layers, (2) the size of the hidden layer(s), and (3) the learning rate used.
\MNAME's results seem quite robust to the choice of hyper-parameters.
\begin{table}[!ht]
\normalsize
\caption{\MNAME's results vary little with different hyper-parameters. Lag-time=5. Noise level = 0.1.
L: number of layers, D: size of hidden layer, R: learning rate (D1: 128, D2: 256, D3: 512, R1: 1E-3, R2: 3E-4, R3: 1E-4)}\label{tbl:arch}
\centering
\setlength{\tabcolsep}{2pt}
\begin{tabular}{lcccccc}
    \toprule
    Simulation & \textbf{L1-D2-R2} & L1-D1-R2 & L1-D3-R2 & L2-D2-R2 & L1-D2-R1 & L1-D2-R3 \\
    \midrule
    \textbf{Random-walk} & $0.97{\pm}0.06$  & $0.96{\pm}0.06$ & $0.96{\pm}0.06$ & $0.96{\pm}0.06$ & $0.92{\pm}0.09$ & $0.97{\pm}0.06$ \\
    \textbf{Sigmoid}     & $0.92{\pm}0.00$  & $0.92{\pm}0.09$ & $0.91{\pm}0.08$ & $0.94{\pm}0.03$ & $0.92{\pm}0.00$ & $0.92{\pm}0.00$
\end{tabular}
\end{table}

\subsection{Results on ADNI Data}
\begin{figure}[!h]
\centering
\hfill
\subfloat[\small{Only ROI volumes}\label{fig:adni_06var}]{\includegraphics[width=0.27\textwidth]{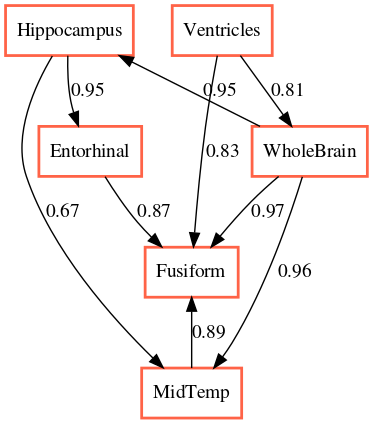}}
\hfill
\subfloat[\small{ROI volumes \& cognitive tests}\label{fig:adni_09var}]{\includegraphics[width=0.33\textwidth]{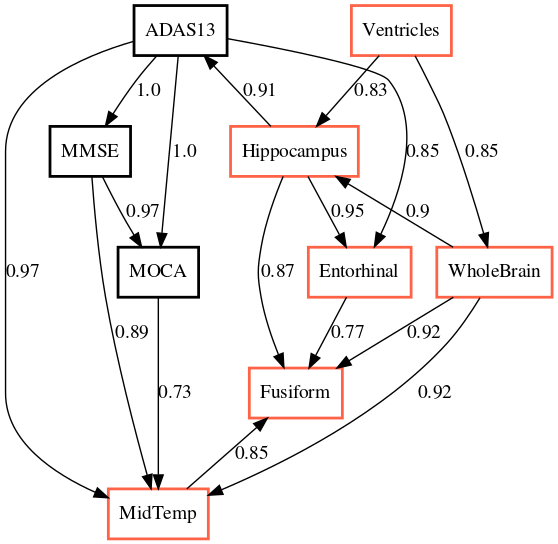}}
\hfill
\subfloat[\small{Extended set of variables}\label{fig:adni_12var}]{\includegraphics[width=0.36\textwidth]{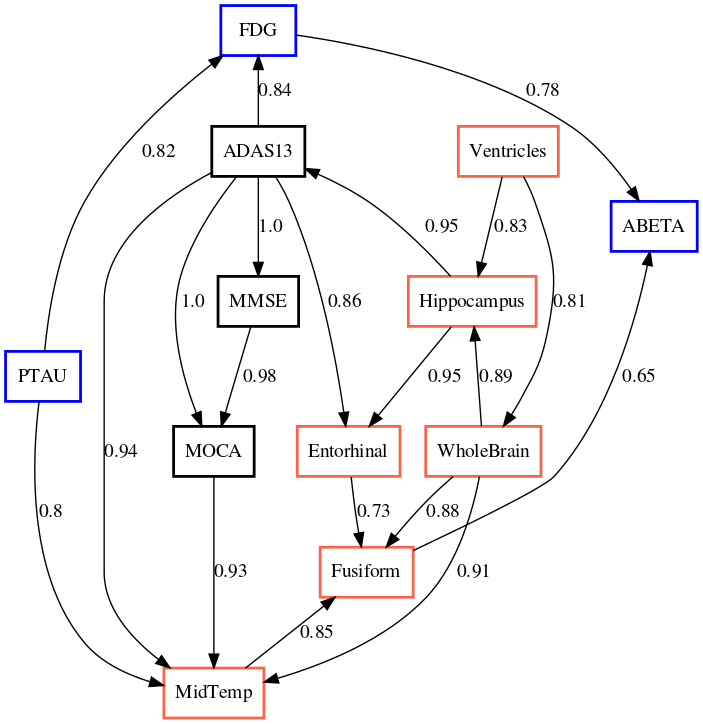}}
\hfill
\caption{\MNAME's predicted interaction of ADNI biomarkers.
ROI volumes are in red, cognitive tests are in black, and the rest are in blue.
ABETA: amyloid beta, PTAU: phosphorylated tau.
Edge weights are frequencies at which edges were detected in multiple runs.}\label{fig:adni}
\end{figure}

The output of applying \MNAME~to different sets of ADNI biomarkers are shown in Fig~\ref{fig:adni}.
Edge weights denote the frequencies at which edges were detected in multiple runs.
Most of the edges are consistently detected across different runs with the exception of ``Hippocampus $\rightarrow$ MidTemp'' (67\%, Fig~\ref{fig:adni_06var}) and ``Fusiform $\rightarrow$ ABETA'' (65\%, Fig~\ref{fig:adni_12var}).
Although \MNAME's neural forecasting model assumes MCAR and missingness in ADNI data may be more adverse than that, \MNAME's result seems promising.
There is a high degree of agreement between the 3 graphs which all show the ``Ventricle'' is a root in the causal graph and ``Fusiform'' is at the end of the chain, which is aligned with prior work~\cite{nestor2008ventricular}.
The presence of the edge ``Hippocampus $\rightarrow$ Entorhinal'' is also consistent with literature~\cite{krumm2016cortical, planche2022structural}.
In comparison, baselines' outputs are less interpretable (Fig~\ref{fig:adni_baselines_main}; more results are in Appendix~\ref{app:adni}).
The outputs of DYNOTEARS and linear Granger contain hardly any edge between ROI volumes while the outputs of cLSTM and eSRU have bidirectional edges.

One potential issue with \MNAME's output is that some cognitive scores are causal parents to some ROI volumes.
This might be due to the no hidden confounder assumption being violated.
Another reason might be measurement noise.
For example, based on our understanding of Alzheimer's dynamics, changes in volumetric measurements should cause cognitive decline, and therefore atrophy in MRI biomarkers should precede changes in clinical scores.
However, initial volumetric changes may be too small to be captured in MRI images and this may affect the inferred graph.

\begin{figure}[!h]
\centering
\includegraphics[width=.95\linewidth]{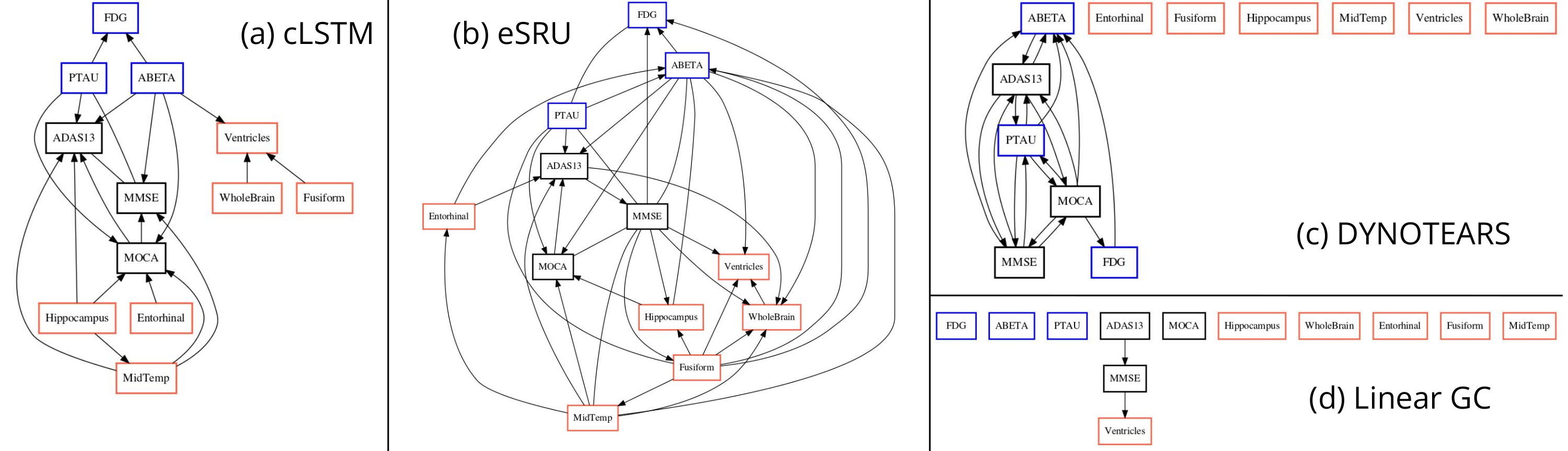}
\caption{Baseline approaches' predicted interaction of ADNI biomarkers.
}\label{fig:adni_baselines_main}
\end{figure}

\section{Discussion}\label{sec:discussion}
Longitudinal studies, in which multiple subjects are sparsely observed for a limited number of times, are particularly common in population health applications.
Longitudinal studies often track many variables, which are likely governed by nonlinear dynamics that might have subject-specific idiosyncrasies.
Yet, longitudinal studies are not amenable to the popular Granger causality (GC) analysis, since GC was developed to analyze a single multivariate densely sampled timeseries.
Furthermore, real-world longitudinal data often suffer from widespread missingness.
We developed \MNAME~which combines the GC framework with a machine learning based prediction model to address the need for a method to find causal relations of numerous variables in longitudinal multi-modal medical imaging studies.
\MNAME~treats subjects as independent samples and uses average prediction accuracy on hold-out subjects to test for causal relations.

\MNAME~exploits a single multi-task neural network trained with input feature dropout to efficiently probe links.
\MNAME~places no restriction on the design of the neural network predictor.
This flexibility allows future extensions of our work.
For example, Transformers~\citep{vaswani2017attention} or Neural ODEs~\citep{chen2018neural} can be used instead of RNN.

Although we showed \MNAME~working well in many settings (varying lag-times, noise levels, and missingness degree), there are some questions remaining that need further investigation.
Even though the ADNI dataset is small by machine learning standard, many scientific datasets are even smaller so characterizing the performance on small datasets would be interesting. 
In addition, it would be interesting to use \MNAME~to analyze more medical imaging datasets~\citep{gutierrez2022connectomic,basaia2022neurogenetic}. %
We focused on continuous variables since they are the most common but extending \MNAME~to discrete variables by adopting techniques in~\citep{peters2010identifying,cai2018causal,huang2018generalized} would make \MNAME~analysis applicable to more longitudinal studies.
Furthermore, studying \MNAME's behaviors under missingness other than MCAR is important despite \MNAME~outputting plausible graphs on real-world data (ADNI).
Missing at random (MAR; missingness probabilities depend on the values of the observed features) and missing not at random (MNAR; missingness depends on the values of the ,unobserved features) are different from MCAR, and MNAR in particular may require different treatments.
Besides, \MNAME~assumes that there is no feedback, hidden confounder, or instantaneous effect.
Thus, before applying \MNAME, it is critical to verify whether these assumptions are reasonable to ensure that causal relations inferred by \MNAME~are valid.
The third assumption in particular requires that the sampling resolution is high enough to capture transient changes (e.g.~impulses) or temporal orderings between causal pairs with short time lags.
Although causal discovery when some assumptions are violated has been studied in the past (for example, presence of hidden confounders~\citep{spirtes2000causation} or presence of instantaneous effects~\citep{danks2013learning}), extending these techniques to longitudinal studies is still an open question.
We leave these questions for future work.

Similar to other causal discovery methods, \MNAME~can offer valuable insights into relationships between variables.
However, intentional or unintentional misuse may lead to harmful consequences.
Causal discovery methods, such as \MNAME, identify potential causal links based on statistical patterns in the data, so biased data collection may lead to misleading findings.
These flawed findings may further lead to misguided medical treatments or clinical decisions that can harm vulnerable populations.
Moreover, violations of modeling assumptions and measurement errors can contribute to mistakes in the inferred graphs.
Thus, it is important to ensure unbiased data collection as well as to corroborate any identified causal links using further experimentation and follow-up study designs to mitigate potential risks when applying causal discovery.

\acks{This research was supported by NIH grants R01LM012719, R01AG053949, the NSF NeuroNex grant 1707312, and the NSF CAREER 1748377 grant (MS).}

\ethics{The work follows appropriate ethical standards in conducting research and writing the manuscript, following all applicable laws and regulations regarding treatment of animals or human subjects.}

\coi{We declare we don't have conflicts of interest.}

\data{Data collection and sharing for this project was funded by the Alzheimer's Disease Neuroimaging Initiative (ADNI) (National Institutes of Health Grant U01 AG024904) and DOD ADNI (Department of Defense award number W81XWH-12-2-0012).
ADNI is funded by the National Institute on Aging, the National Institute of Biomedical Imaging and Bioengineering, and through generous contributions from the following: AbbVie, Alzheimer's Association; Alzheimer's Drug Discovery Foundation; Araclon Biotech; BioClinica, Inc.; Biogen; Bristol-Myers Squibb Company; CereSpir, Inc.; Cogstate; Eisai Inc.; Elan Pharmaceuticals, Inc.; Eli Lilly and Company; EuroImmun; F. Hoffmann-La Roche Ltd and its affiliated company Genentech, Inc.; Fujirebio; GE Healthcare; IXICO Ltd.; Janssen Alzheimer Immunotherapy Research \& Development, LLC.; Johnson \& Johnson Pharmaceutical Research \& Development LLC.; Lumosity; Lundbeck; Merck \& Co., Inc.; Meso Scale Diagnostics, LLC.; NeuroRx Research; Neurotrack Technologies; Novartis Pharmaceuticals Corporation; Pfizer Inc.; Piramal Imaging; Servier; Takeda Pharmaceutical Company; and Transition Therapeutics.
The Canadian Institutes of Health Research is providing funds to support ADNI clinical sites in Canada.
Private sector contributions are facilitated by the Foundation for the National Institutes of Health (www.fnih.org).
The grantee organization is the Northern California Institute for Research and Education, and the study is coordinated by the Alzheimer's Therapeutic Research Institute at the University of Southern California.
ADNI data are disseminated by the Laboratory for Neuro Imaging at the University of Southern California.}

\bibliography{cause}

\clearpage
\appendix
\section{Additional Details about \MNAME}
\subsection{Granger Causality MSE Test}\label{app:gc_mse}
Let $\mathbf{X}_t$ and $\mathbf{X}^t= \{\mathbf{X}_{0},\ldots,\mathbf{X}_{t-1}\}$ denote a time-varying random variables and its history.
Let $\mathbf{\Omega}^t = \mathbf{X}^t \cup \mathbf{Y}^t \cup \dots$ be the union of all historical variables available at time $t$.
The general GC hypothesis is:
$Y$ causing $X$ $\Rightarrow$ there exists some event $A$ and $t \in [T]$ such that
\[ \Pr(\mathbf{X}_{t}\in A | \mathbf{\Omega}^{t}) \neq \Pr(\mathbf{X}_{t}\in A | \mathbf{\Omega}^{t} \setminus \mathbf{Y}^{t}) \]
Equivalently, we have:
\begin{align*}
    &\Pr(\mathbf{X}_t | \mathbf{\Omega}^{t}) = \Pr(\mathbf{X}_t | \mathbf{\Omega}^{t} \setminus \mathbf{Y}^{t}), \forall t \in [T] \\
    &\Rightarrow Y \;\text{does not cause}\; X
\end{align*}
In general it is not possible to compare these conditional probabilities based on observed timeseries data.
A practical approach is to compare conditional expectations:
\[ \EE[\mathbf{X}_t | \mathbf{\Omega}^{t}] \stackrel{?}{=} \EE[\mathbf{X}_t | \mathbf{\Omega}^{t} \setminus \mathbf{Y}^{t}]. \]
Note that conditional expectations can be infeasible to compare, so we often need further assumptions.
One observation is that the conditional expectation is the optimal estimator that minimizes the mean square error (MSE).
This gives rise to a GC test that compares the MSE of least-squares predictors. In this approach, we conclude that ``$Y$ causes $X$'' if:
\begin{align}
\MSE(\mathbf{X}_t, \EE[\mathbf{X}_t | \mathbf{\Omega}^{t}]) < \MSE(\mathbf{X}_t, \EE[\mathbf{X}_t | \mathbf{\Omega}^{t} \setminus \mathbf{Y}^{t}]) \label{eq:GC-MSE}
\end{align}
Note that, in above equation the right hand side is larger than or equal to the left hand side because, in general, the least-square loss will be smaller with more history.

In practice, the implementation of the GC MSE test often relies on two more assumptions.
The first is the stationarity assumption.
That is, we suppose $\EE[\mathbf{X}_t | \mathbf{\Omega}^{t}]$ and $\EE[\mathbf{X}_t | \mathbf{\Omega}^{t} \setminus \mathbf{Y}^{t}]$ are independent of $t$.
Second, we assume the stochastic processes are Markovian and thus a finite history (often just the prior timepoint) is sufficient for making the least-square forecast. In our framework, the Markovian assumption can be relaxed because the RNN forecast model can digest all available history.

\subsection{\MNAME's First Heuristic Justification}\label{app:first_heuristic}
In this section, we will provide some justification for our post-processing heuristic that removes one of the arrows of a bi-directional edge.
We will consider a scenario where only one previous timepoint is given to predict the future timepoint.
Furthermore, we will suppose that we have infinite data and infinite capacity models that allow us to estimate the unknown parameters and conditional expectations exactly.
\begin{align}
    \shortintertext{We will assume following data generation process (Fig~\ref{fig:temporal_causal_diagram}) with constants $a,b,c > 0$ and independent and identically-distributed, zero-mean additive errors $\epsilon_{X,t}, \epsilon_{Y,t}$ }
    X_t     &:= a X_{t-1} + b X_{t-2} + \epsilon_{X,t} \label{eq:x_gen} \\
    Y_t     &:= c X_{t-1} + \epsilon_{Y,t} \label{eq:y_gen} \\
    \EE[\epsilon_{X,t}^2] &= \sigma^2_{X} > 0 ;\;\;\;\; \EE[\epsilon_{Y,t}^2] =\sigma^2_{Y} > 0 \\
    \shortintertext{Back up 1 step in time}
    X_{t-1} &= a X_{t-2} + b X_{t-3} + \epsilon_{X,t-1} \label{eq:x_gen_2} \\
    Y_{t-1} &= c X_{t-2} + \epsilon_{Y,t-1} \label{eq:y_gen_2}
\end{align}

\begin{figure}[!h]
    \centering
    \begin{tikzpicture}[> = stealth, shorten > = 1pt, auto, node distance = 2.0cm, thick]
        \tikzstyle{every state}=[draw=black, thick, fill=white, minimum size=1.2cm]
        \node[state] (X0) {$X_t$};
        \node[state] (X1) [left of=X0]{$X_{t-1}$};
        \node[state] (X2) [left of=X1]{$X_{t-2}$};
        \node[state] (X3) [left of=X2]{$X_{t-3}$};
        \node[state] (Y0) [below of=X0] {$Y_t$};
        \node[state] (Y1) [left of=Y0]{$Y_{t-1}$};
        \node[state] (Y2) [left of=Y1]{$Y_{t-2}$};
        \node[state] (Y3) [left of=Y2]{$Y_{t-3}$};

        \path[->, line width=1] (X1) edge node {} (X0);
        \path[->, line width=1, out=45, in=135] (X2) edge node {} (X0);

        \path[->, line width=1] (X2) edge node {} (X1);
        \path[->, line width=1, out=45, in=135] (X3) edge node {} (X1);

        \path[->, line width=1] (X3) edge node {} (X2);

        \path[->, line width=1] (X3) edge node {} (Y2);
        \path[->, line width=1] (X2) edge node {} (Y1);
        \path[->, line width=1] (X1) edge node {} (Y0);

    \end{tikzpicture}
    \caption{{\bf Data generation.} $X$ is an order-2 auto-regressive process and $X$ causes $Y$ ($X{\rightarrow}Y$).
    }\label{fig:temporal_causal_diagram}
\end{figure}

\begin{align}
    \shortintertext{Computing $\MSE(X_t, \EE[X_t|X_{t-1},Y_{t-1}])$}
    X_{t-2} &= \frac{1}{c}\Big(Y_{t-1} - \epsilon_{Y,t-1}\Big) \label{eq:unk1} &\text{from (\ref{eq:y_gen_2})} \\
    X_t     &= a X_{t-1} + \frac{b}{c}\Big(Y_{t-1} - \epsilon_{Y,t-1}\Big) + \epsilon_{X,t} &\text{from (\ref{eq:x_gen}) and (\ref{eq:unk1})} \nonumber\\
    &= a X_{t-1} + \frac{b}{c}Y_{t-1} - \frac{b}{c}\epsilon_{Y,t-1} + \epsilon_{X,t} \label{eq:unk2}
\end{align}

\begin{align}
    \shortintertext{Since $\epsilon_{Y,t-1}$, $\epsilon_{X,t}$ are independent of $X_{t-1}$, $Y_{t-1}$, from (\ref{eq:unk2}):}
    \Rightarrow \EE[X_t|X_{t-1},Y_{t-1}] &= a X_{t-1} + \frac{b}{c}Y_{t-1} \\
    \Rightarrow X_t - \EE[X_t|X_{t-1},Y_{t-1}] &= -\frac{b}{c}\epsilon_{Y,t-1} + \epsilon_{X,t} \\
    \MSE(X_t, \EE[X_t|X_{t-1},Y_{t-1}]) &= \EE[(X_t - \EE[X_t|X_{t-1},Y_{t-1}])^2] \nonumber\\
    &= \EE\Big[\Big(-\frac{b}{c}\epsilon_{Y,t-1} + \epsilon_{X,t}\Big)^2\Big]
    = \frac{b^2}{c^2}\sigma^2_Y + \sigma^2_X
\end{align}

\begin{align}
    \shortintertext{Computing $\MSE(X_t, \EE[X_t|X_{t-1}])$}
    X_{t-2} &= \frac{1}{a} X_{t-1} - \frac{b}{a} X_{t-3} - \frac{1}{a} \epsilon_{X,t-1} \label{eq:unk3} \quad\text{from (\ref{eq:x_gen_2})} \\
    X_t     &= a X_{t-1} + b\Big(\frac{1}{a} X_{t-1} - \frac{b}{a} X_{t-3} - \frac{1}{a} \epsilon_{X,t-1}\Big)
    + \epsilon_{X,t} \quad\text{from (\ref{eq:x_gen}) and (\ref{eq:unk3})} \nonumber\\
    &= \Big(a+\frac{b}{a}\Big)X_{t-1} - \frac{b^2}{a} X_{t-3} - \frac{b}{a} \epsilon_{X,t-1} + \epsilon_{X,t}  \label{eq:unk4}
\end{align}
\begin{align}
    \shortintertext{Since $\epsilon_{X,t-1}$, $\epsilon_{X,t}$ are independent of $X_{t-1}$, and assuming the correlation between $X_{t-1}$ and $X_{t-3}$ is weak, from (\ref{eq:unk4}):}
    \EE[X_t|X_{t-1}] &\approx \Big(a+\frac{b}{a}\Big)X_{t-1} \\
    X_t - \EE[X_t|X_{t-1}] &\approx -\frac{b^2}{a} X_{t-3} - \frac{b}{a} \epsilon_{X,t-1} + \epsilon_{X,t}
\end{align}

\begin{align}
    \MSE(X_t, \EE[X_t|X_{t-1}]) &\approx \EE[(X_t - \EE[X_t|X_{t-1}])^2] \nonumber\\
    &\approx \EE\Big[\Big(-\frac{b^2}{a} X_{t-3} - \frac{b}{a} \epsilon_{X,t-1} + \epsilon_{X,t}\Big)^2\Big] \\
    &\approx \frac{b^4}{a^2} \EE[X^2_{t-3}] + \frac{b^2}{a^2} \sigma^2_X  + \sigma^2_X  \quad\text{(independent)} \\ 
    \shortintertext{Estimating $\Delta\MSE(X|Y)$}
    \Delta\MSE(X|Y) &= \MSE(X_t, \EE[X_t|X_{t-1}]) - \MSE(X_t, \EE[X_t|X_{t-1},Y_{t-1}]) \\
    &\approx \Big(\frac{b^4}{a^2} \EE[X^2_{t-3}] + \frac{b^2}{a^2} \sigma^2_X  + \sigma^2_X\Big) - \Big(\frac{b^2}{c^2}\sigma^2_Y + \sigma^2_X\Big) \\
    &\approx\frac{b^4}{a^2} \EE[X^2_{t-3}] + \frac{b^2}{a^2} \sigma^2_X  - \frac{b^2}{c^2}\sigma^2_Y
\end{align}

Computing $\MSE(Y_t, \EE[Y_t|X_{t-1},Y_{t-1}])$
\begin{align}
    &\Rightarrow \EE[Y_t|X_{t-1},Y_{t-1}] = c X_{t-1} \quad\text{from (\ref{eq:y_gen})} \\
    &\Rightarrow Y_t - \EE[Y_t|X_{t-1},Y_{t-1}] = \epsilon_{Y,t} \quad\text{from (\ref{eq:y_gen})} \\
    &\Rightarrow \MSE(Y_t, \EE[Y_t|X_{t-1},Y_{t-1}]) \nonumber\\
    &= \EE[(Y_t - \EE[Y_t|X_{t-1},Y_{t-1}])^2] = \EE[(\epsilon_{Y,t})^2] = \sigma^2_Y
\end{align}

Estimating $\MSE(Y_t, \EE[Y_t|Y_{t-1}])$
\begin{align}
    \shortintertext{From (\ref{eq:x_gen_2}) and (\ref{eq:unk1}):}
    X_{t-1} &= \frac{a}{c}\Big(Y_{t-1} - \epsilon_{Y,t-1}\Big) + b X_{t-3} + \epsilon_{X,t-1} \label{eq:unk5}
    \shortintertext{From (\ref{eq:y_gen}) and (\ref{eq:unk5}):}
    Y_t     &= c \Big[\frac{a}{c}\Big(Y_{t-1} - \epsilon_{Y,t-1}\Big) + b X_{t-3} + \epsilon_{X,t-1}\Big] + \epsilon_{Y,t} \nonumber\\
    &= a Y_{t-1} - a \epsilon_{Y,t-1} + bc X_{t-3} + c \epsilon_{X,t-1} + \epsilon_{Y,t} \label{eq:unk6}
\end{align}

Since $\epsilon_{X,t-1}$, $\epsilon_{Y,t}$ are independent of $Y_{t-1}$, and assuming the correlation between $Y_{t-1}$ and $X_{t-3}$ is weak, from (\ref{eq:unk6}):
\begin{align}
    \Rightarrow \EE[Y_t|Y_{t-1}] &\approx a Y_{t-1} \\
    \Rightarrow Y_t - \EE[Y_t|Y_{t-1}] &\approx - a \epsilon_{Y,t-1} + bc X_{t-3} + c \epsilon_{X,t-1} + \epsilon_{Y,t}
\end{align}

\begin{align}
    &\Rightarrow \MSE(Y_t, \EE[Y_t|Y_{t-1}]) = \EE[(Y_t - \EE[Y_t|Y_{t-1}])^2] \nonumber\\
    &\approx \EE[(- a \epsilon_{Y,t-1} + bc X_{t-3} + c \epsilon_{X,t-1} + \epsilon_{Y,t})^2] \\
    &\approx a^2 \sigma^2_Y + b^2c^2 \EE[X^2_{t-3}] + c \sigma^2_X + \sigma^2_Y \quad\text{(independent)} 
\end{align}

Estimating $\Delta\MSE(Y|X)$
\begin{align}
    \Delta\MSE(Y|X) &= \MSE(Y_t, \EE[Y_t|Y_{t-1}]) - \MSE(Y_t, \EE[Y_t|X_{t-1},Y_{t-1}]) \\
    &\approx (a^2 \sigma^2_Y + b^2c^2 \EE[X^2_{t-3}] + c^2 \sigma^2_X + \sigma^2_Y) - (\sigma^2_Y ) \\
    &\approx a^2 \sigma^2_Y + b^2c^2 \EE[X^2_{t-3}] + c^2 \sigma^2_X
\end{align}

Since $\Delta\MSE(Y|X) > 0$, edge $X{\rightarrow}Y$ is always detected.
Edge $Y{\rightarrow}X$ is (falsely) detected if $\Delta\MSE(X|Y) > 0$.
The heuristic needs to remove this falsely detected edge.
We can show that if false detection happens, our heuristic of comparing $\Delta\MSE(Y|X)$ and $\Delta\MSE(X|Y)$ (which is based on the t-statistics) will show the true direction.
In other words, let's show that if $\Delta\MSE(X|Y) > 0$, then $\Delta\MSE(Y|X) - \Delta\MSE(X|Y) > 0$. 

\begin{align}
    &\Delta\MSE(X|Y) > 0
    \Leftrightarrow \frac{b^4}{a^2} \EE[X^2_{t-3}] + \frac{b^2}{a^2} \sigma^2_X  - \frac{b^2}{c^2}\sigma^2_Y > 0
    \Leftrightarrow b^2 \EE[X^2_{t-3}] > \frac{a^2}{c^2}\sigma^2_Y - \sigma^2_X \label{eq:unk7}\\
    &\Delta\MSE(Y|X) - \Delta\MSE(X|Y) \nonumber\\
    &\approx (a^2 \sigma^2_Y + b^2c^2 \EE[X^2_{t-3}] + c^2 \sigma^2_X) - \Big(\frac{b^4}{a^2} \EE[X^2_{t-3}] + \frac{b^2}{a^2} \sigma^2_X  - \frac{b^2}{c^2}\sigma^2_Y\Big) \\
    &\approx a^2 \sigma^2_Y + b^2\EE[X^2_{t-3}]\Big(c^2 - \frac{b^2}{a^2}\Big) + c^2 \sigma^2_X - \frac{b^2}{a^2} \sigma^2_X  + \frac{b^2}{c^2}\sigma^2_Y \label{eq:unk8}
\end{align}

Let $\mathsf{Diff}=\Delta\MSE(Y|X) - \Delta\MSE(X|Y)$, from (\ref{eq:unk7}) and (\ref{eq:unk8})
\begin{align}
    \mathsf{Diff} &> a^2 \sigma^2_Y + \Big(\frac{a^2}{c^2}\sigma^2_Y - \sigma^2_X\Big)\Big(c^2 - \frac{b^2}{a^2}\Big) &+ c^2 \sigma^2_X - \frac{b^2}{a^2} \sigma^2_X  + \frac{b^2}{c^2}\sigma^2_Y  \\
    \Leftrightarrow \mathsf{Diff} &> a^2 \sigma^2_Y + a^2\sigma^2_Y - c^2 \sigma^2_X - \frac{b^2}{c^2}\sigma^2_Y + \frac{b^2}{a^2} \sigma^2_X &+ c^2 \sigma^2_X - \frac{b^2}{a^2} \sigma^2_X  + \frac{b^2}{c^2}\sigma^2_Y \\
    \Leftrightarrow \mathsf{Diff} &> 2a^2 \sigma^2_Y > 0 \;\;\;\;\text{(Q.E.D)}
\end{align}

\section{Pseudo-code for Data Generation}\label{app:data_gen}
Algorithm~\ref{algo:data} shows the steps to generate data from a given causal graph $G$.
First, the edge weights are sampled.
Then, for each subject, the timeseries are generated in topological order.
If a node has no parent, i.e., if it is a source node, its timeseries is specified by the sample path $f$ (Gaussian random-walk or sigmoid).
The random-walk function is a conventional choice while the sigmoid function yields trajectories that mimic the evolution of many real-world dynamical systems~\citep{jack2013tracking}.
A non-source node's timeseries is the weighted sum of the lagged version of its parents' timeseries.
Next, Gaussian measurement noise with standard deviation $\sigma$ is added to the timeseries.
Finally, a discrete set of timepoints within a randomly-chosen observation window are extracted, mimicking a real-world longitudinal study.
\begin{itemize}
     \item Gaussian random-walk: $f(t) = \sum_{i=0}^{t} \mathcal{N}(0, 1)$
     \item Sigmoid: $f(t) = \frac{A}{1+e^{-k(t-t_0)}};\quad A\sim\mathsf{Unif}(1, 2),\; t_0\sim\mathsf{Unif}(40, 60),\; k\sim\mathsf{Unif}(0.1, 0.3)$
\end{itemize}
For random-walk timeseries, the noise variance $\sigma$ is either 0.1 or 1.0 (smaller $\sigma$ has no visible effect).
For sigmoid timeseries, $\sigma = 0.001, 0.01, 0.1, 1.0$.
Since measurement noise can (1) induce spurious causality between unrelated variables and (2) suppress true causality~\citep{newbold1978feedback,glymour2019review}, it is important to benchmark across different levels of measurement noise.
For each set of parameters ($f$, lag-time $L$, and $\sigma$), we generated 5 different randomized datasets so as to estimate the standard error of the performance metrics.

\begin{algorithm}[H]
\caption{Data Generation}\label{algo:data}
\begin{algorithmic}
\State \textbf{In}: Causal graph $G$, sample path $f$, number of subjects $n$, number of timepoints $m$, Lag-time $L$, measurement noise magnitude $\sigma$
\State \textbf{Out}: Dataset $D=(X_1,\dots,X_n)$
\State \textsc{Sample edge weight}
\State \hspace{0.0cm} \textbf{For each} $u{\rightarrow}v\in G$
\State \hspace{0.3cm}     $s_{uv}\sim\mathsf{Rademacher}()$;
\State \hspace{0.3cm}     $m_{uv}\sim\mathsf{Unif}(0.5, 1)$;
\State \hspace{0.3cm}     $w_{uv} = s_{uv}*m_{uv}$
\State \textsc{Sample observation series of a subject}
\State \hspace{0.0cm} \textbf{For each} subject $i$
\State \hspace{0.3cm}     \textbf{For each} node $v$
\State \hspace{0.6cm}         $b \sim \mathsf{Unif}(-0.5, 0.5)$ (bias term)
\State \hspace{0.6cm}         \textbf{If} $v$ has no parent
\State \hspace{0.9cm}             $s_v[t] = b + f[t]$     (time $t \in [0, 100]$)
\State \hspace{0.6cm}         \textbf{Else}
\State \hspace{0.9cm}             $s_v[t] = b + \sum_{u, (u, v)\in G} w_{uv} * s_u[t - L]$
\State \hspace{0.6cm}         $s_v[t] = s_v[t] + \mathcal{N}(0,\sigma)$     (measurement noise)
\State \hspace{0.3cm}     $S_i = \{s_u, u\in G\}$     (subject data)
\State \hspace{0.3cm}     $t_{\text{start}}\sim\mathsf{Unif}(30, 70)$;
\State \hspace{0.3cm}     $t_{\text{end}} = t_{\text{start}} + m$;
\State \hspace{0.3cm}     $X_i = S_i[t_{\text{start}}: t_{\text{start}}]$ (extract timepoints within window)
\end{algorithmic}
\end{algorithm}

In addition to the linear structural causal model (SCM) shown in Algorithm~\ref{algo:data}, we also experimented with 2 non-linear SCMs.
One SCM takes a polynomial form (Eq.~\ref{eq:poly}) while the other takes a trigonometric form (Eq.~\ref{eq:trig}).
\begin{align}
    c_{uv} &\sim \mathsf{Unif}(-1, 1) \quad\quad\quad p_{uv} \sim \mathsf{Unif}(0.1, 1) \nonumber\\
    s_v[t] &= \mathsf{Unif}(-0.5, 0.5) + \mathcal{N}(0,\sigma) + \sum_{u, (u, v)\in G} w_{uv} * |s_u[t - L] + c_{uv}|^{p_{uv}} \label{eq:poly}\\
    s_v[t] &= \mathsf{Unif}(-0.5, 0.5) + \mathcal{N}(0,\sigma) + \sum_{u, (u, v)\in G} w_{uv} * \sin(s_u[t - L] + c_{uv}) \label{eq:trig}
\end{align}

\section{Additional Simulation Results}
\subsection{More Repetitions}\label{app:more_reps}
\begin{figure}[!h]
\centering
\subfloat[Lag-time (L) = 5]{\includegraphics[width=.8\linewidth]{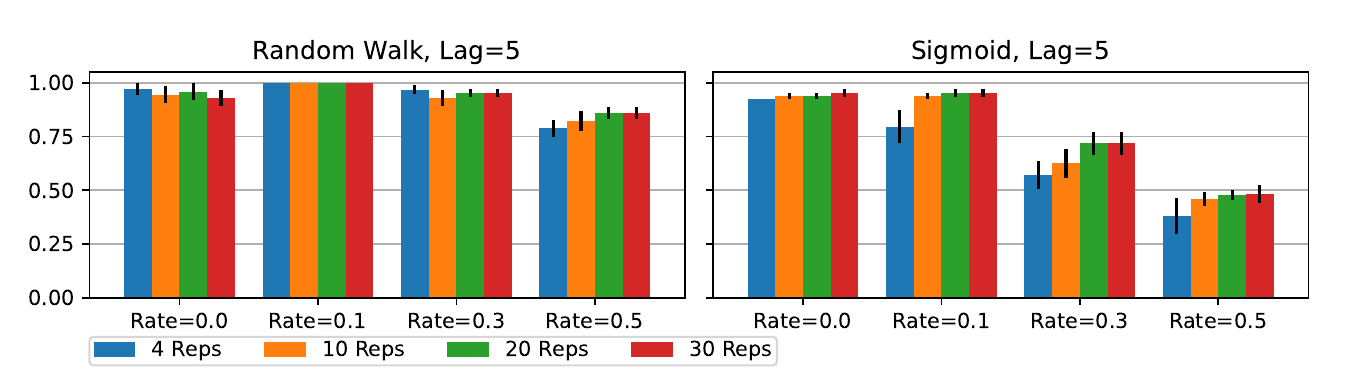}}
\vfill
\subfloat[Lag-time (L) = 7]{\includegraphics[width=.8\linewidth]{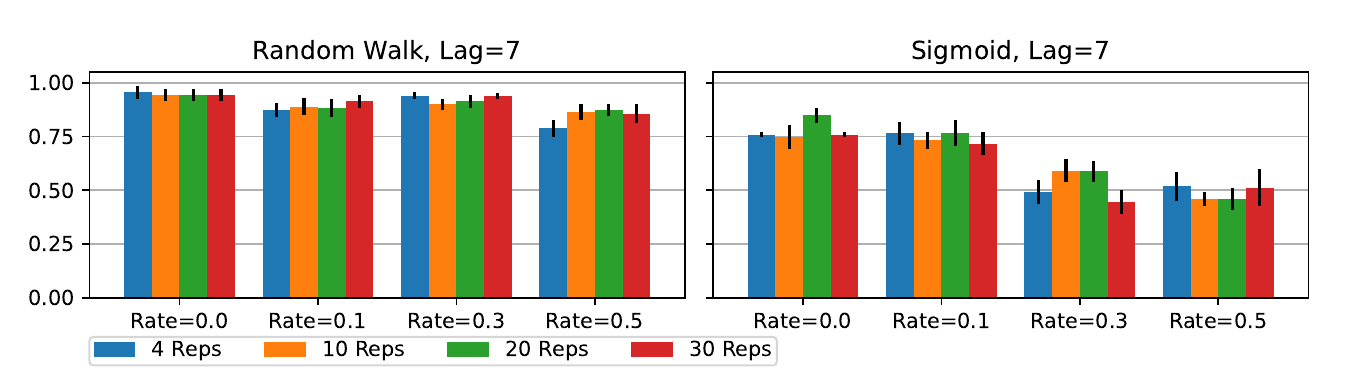}}
\caption{More repetitions of cross-validation lead to slightly better result at the expense of running time.}\label{fig:more_reps}
\end{figure}

\MNAME's results in Section~\ref{ssec:sim_result} were obtained by repeating 5-fold cross-validation for 4 times.
Increasing the number of repetitions leads to higher F-1 score as shown by the trend in Fig~\ref{fig:more_reps}.
However, the gap between 10 repetitions and 4 repetitions is not large enough to justify the extra computational cost of repeating more than 4 times in Section~\ref{ssec:sim_result}.
Another interesting trend in Fig~\ref{fig:more_reps} is that the gap between 4 repetitions and 30 repetitions is more apparent for missing data.
Thus, when the data is noisy (more missing values), more repetitions may yield more accurate results.

\clearpage
\subsection{More Comparisons of \MNAME~against Baselines}\label{app:more_comparisons}
\begin{figure}[!h]
\centering
\includegraphics[width=.95\linewidth]{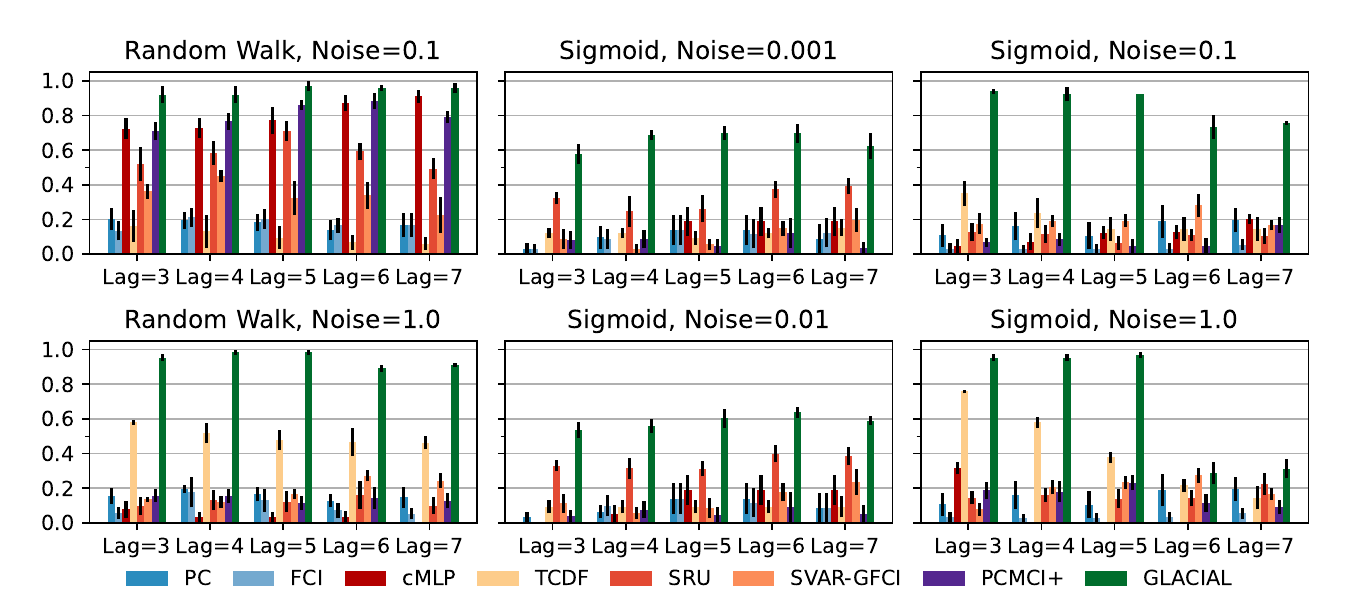}
\caption{Average F1-scores at different lag-time and measurement noise for 7-node graph.
\MNAME~outperforms baselines in most settings of sample path, lag-time, and measurement noise (also see Fig~\ref{fig:main_7n1}).
}\label{fig:extra_7n1}
\end{figure}
\begin{figure}[!h]
\centering
\includegraphics[width=.8\linewidth]{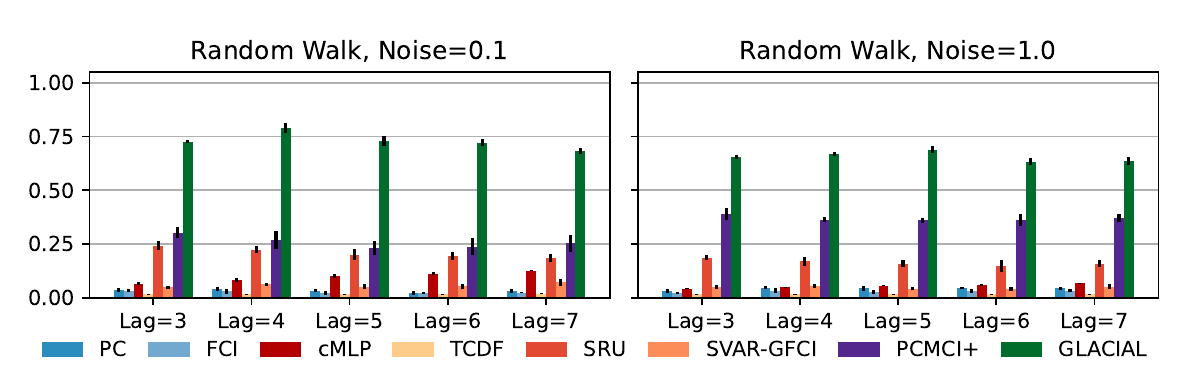}
\caption{Average F1-scores at different lag-time and measurement noise for 39-node graph (Gaussian random-walk).
\MNAME~outperforms baselines in most settings (also see Fig~\ref{fig:main_bro_rtk}).
}\label{fig:extra_bro_rtk}
\end{figure}

\clearpage
\subsection{Non-linear SCM simulations}\label{app:non_linear_scm}

\begin{figure}[!h]
\centering
\includegraphics[width=0.8\linewidth]{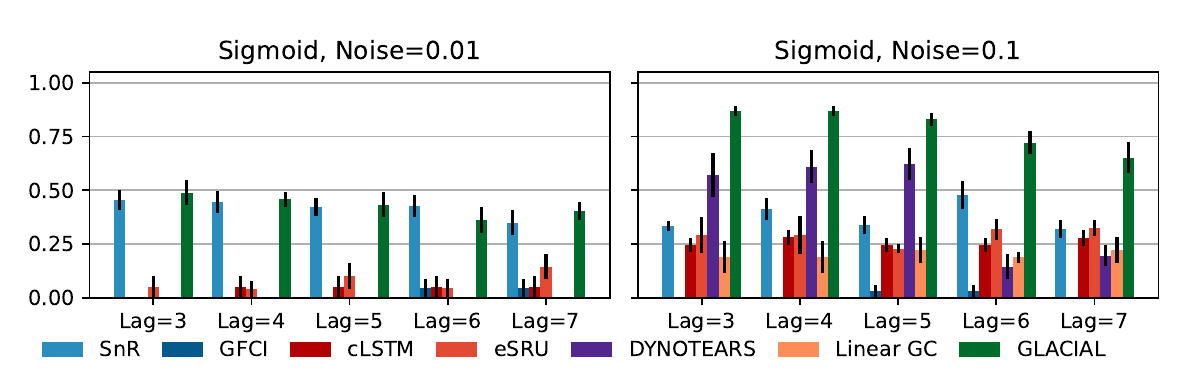}
\caption{F1-scores at different lag-time and measurement noise (polynomial SCM)}\label{fig:scm_poly}
\end{figure}

\begin{figure}[!h]
\centering
\includegraphics[width=0.8\linewidth]{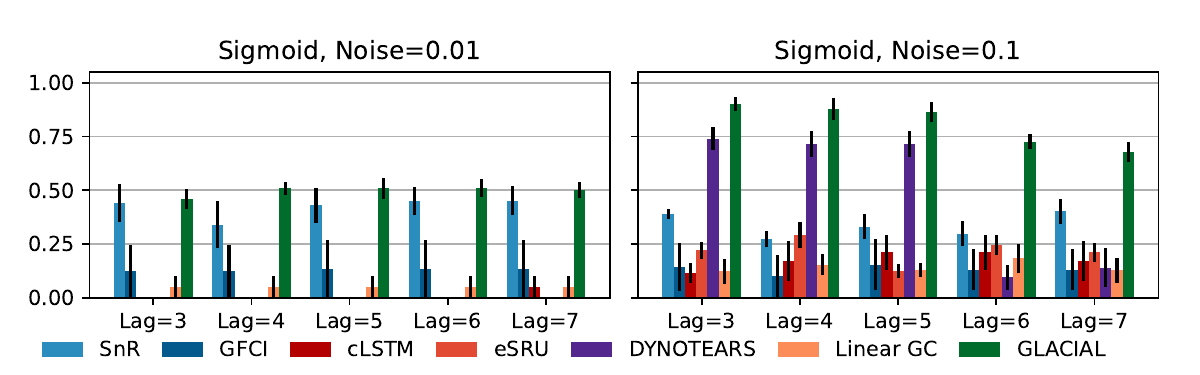}
\caption{F1-scores at different lag-time and measurement noise (trigonometric SCM)}\label{fig:scm_trig}
\end{figure}

\subsection{Constraint-based Baselines with Higher Threshold}\label{app:higher_threshold}
\begin{figure}[!h]
\centering
\includegraphics[width=.95\linewidth]{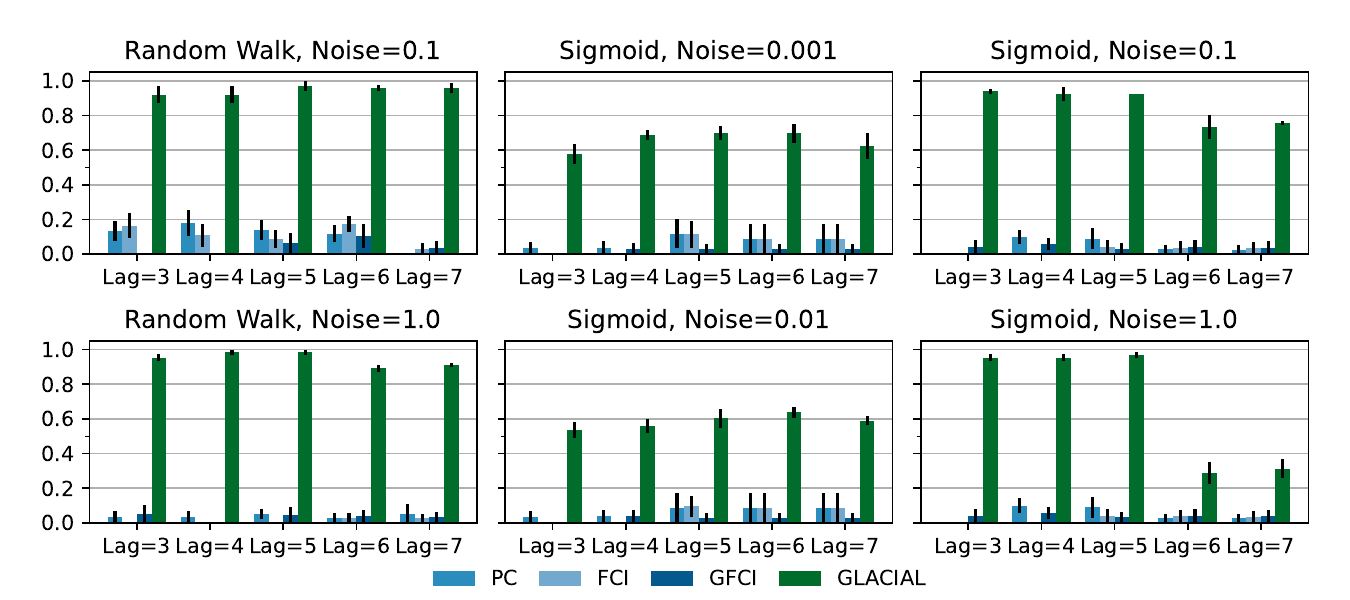}
\caption{Performance of PC, FCI, and GFCI when thresholded at 0.8 (80\%).}\label{fig:th08}
\end{figure}

The PC, FCI, and GFCI baselines are run multiple times using different data bootstraps, resulting in multiple graphs.
To combine the graphs, we only retain edges that appear more than half of the runs.
This procedure is similar to~\citep{shen2020challenges} although they used a more conservative threshold (0.8) in their work.
Fig~\ref{fig:th08} shows the results of the constraint-based baselines when the threshold of 0.8 (80\%) is used.
Compared to the results in Fig~\ref{fig:main_7n1}, this more conservative threshold led to worse performance in the baselines.

\subsection{More Densely Sampled Data}\label{app:more_timepoints}
\begin{figure}[!h]
\centering
\includegraphics[width=.95\linewidth]{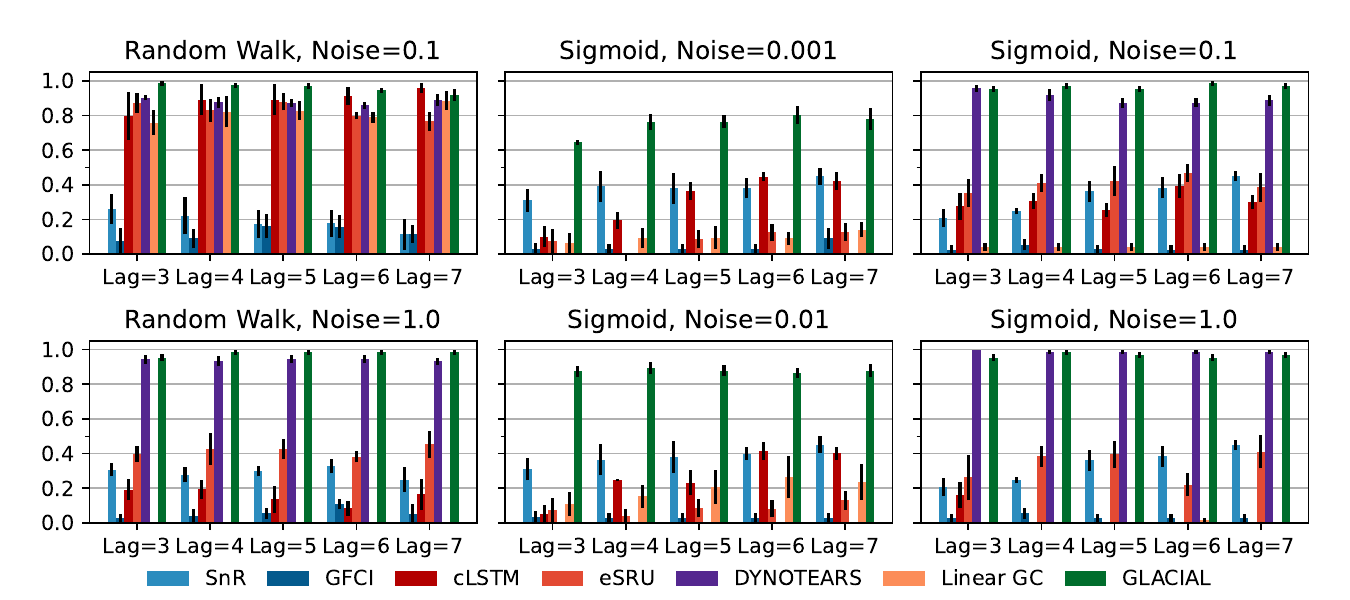}
\caption{Average F1-scores at different settings of sample path, lag-time and measurement noise (7-node graph).
Each subject has 24 timepoints.
\MNAME~outperforms baselines in most settings.
}\label{fig:long_7n1}
\end{figure}

In Section~\ref{ssec:sim_result}, each subject only has 6 timepoints (sparse observations) so linear GC did not work well.
Thus, we investigated a scenario more favorable for linear GC where for when subjects have more timepoints (i.e. 24 timepoints).
With more timepoints, linear GC results improve slightly but are still worse than that of \MNAME~(Fig~\ref{fig:long_7n1}).
Additionally, using GC to estimate one causal graph for each subject could not find the correct graph even with 24 timepoints (hence, result not reported).

\section{Experiment on ADNI Dataset}\label{app:adni}
\subsection{Description of ADNI Variables}
Table~\ref{tbl:adni_variables} shows the ADNI variables used and how they were measured (Data Modality).
These variables are complementary in what they measure.
``ABETA'' and ``PTAU'' measure the level of two proteins in cerebral spinal fluid that are indicative of Alzheimer's disease.
``FDG'' measures brain cells' metabolism while cognitive tests measure performance in various areas such as general cognition, memory, language et cetera.
The quantitative variables derived from structural MRI scans (e.g.\ Hippocampus Volume) is often considered as a proxy of regional brain atrophy, or tissue loss linked to aging and/or neuro-degenerative processes.
\begin{table}[!ht]
\normalsize
\caption{ADNI variables used for causal discovery}\label{tbl:adni_variables}
\centering
\begin{tabular}{lll}
Variable & Description & Data Modality \\
\midrule
ABETA & Amyloid beta & Cerebral spinal fluid \\
FDG & Fluorodeoxyglucose PET & PET imaging \\
PTAU & Phosphorylated tau & Cerebral spinal fluid \\
ADAS13 & ADAS-Cog13 & Cognitive test \\
MMSE & Mini-Mental State Examination & Cognitive test \\
MOCA & Montreal Cognitive Assessment & Cognitive test \\
Entorhinal & Entorhinal cortical volume & MRI imaging \\
Fusiform & Fusiform cortical volume & MRI imaging \\
Hippocampus & Hippocampus volume & MRI imaging \\
MidTemp & Middle temporal cortical volume & MRI imaging \\
Ventricles & Ventricles volume & MRI imaging \\
WholeBrain & Whole brain volume & MRI imaging
\end{tabular}
\end{table}

We normalized the volumetric variables of each subject by dividing the measurements by the subject's intracranial volume, or total head size, which is typically constant in adulthood.
This is a standard normalization done to account for inter-subject variability in head sizes.
FDG is a standardized uptake value ratio computed by dividing the average PET signal in an Alzheimer implicated region of interest to the signal in a control reference region.
The Cerebral Spinal fluid markers correlate with the accumulation of the two Alzheimer's associated pathological proteins in the brain, namely tau tangles and amyloid plaque.

\subsection{Interpretation of ADNI Causal Graphs}
The order of the volumetric variables in Fig~\ref{fig:adni}a,~\ref{fig:adni}b, and~\ref{fig:adni}c are mostly consistent with each other and prior literature on neuroimaging in aging and Alzheimer's disease, where the size of ventricles and whole brain are earliest MRI markers of aging, and Alzheimer's associated atrophy starts at the hippocampus, from where it spreads to cortical areas such as entorhinal and fusiform.The causal chains that appear in all three graphs are:
\begin{itemize}
    \item ``Ventricles'' $\rightarrow$ ``WholeBrain'' $\rightarrow$ ``Hippocampus'' $\rightarrow$ ``Entorhinal'' $\rightarrow$ ``Fusiform''
    \item ``Ventricles'' $\rightarrow$ ``WholeBrain'' $\rightarrow$ ``MidTemp'' $\rightarrow$ ``Fusiform''
\end{itemize}
The ordering of cognitive tests are also consistent across all graphs.
When we examine the ordering of variables from different data modalities, the causal chain of ``Hippocampus'' $\rightarrow$ ``ADAS13'' $\rightarrow$ ``MMSE'' $\rightarrow$ ``MOCA'' is quite interesting.
This implies that the atrophy of the hippocampus, a brain region that plays a central role in memory and learning, leads to worse performance in tasks measured by cognitive tests. The relationship between MMSE and ADAS13 is surprising and deserves follow-up investigation, because classically MMSE is thought of as the earlier marker of cognitive impairment and ADAS13 is a measure of symptoms that appear later in Alzheimer's disease. That said, to our knowledge, we are not aware of a study that examines the relationships between the temporal dynamics of these test scores. Our results indicate that changes in ADAS13 might foreshadow changes in MMSE.

\begin{figure}[!h]
\centering
\includegraphics[width=.95\linewidth]{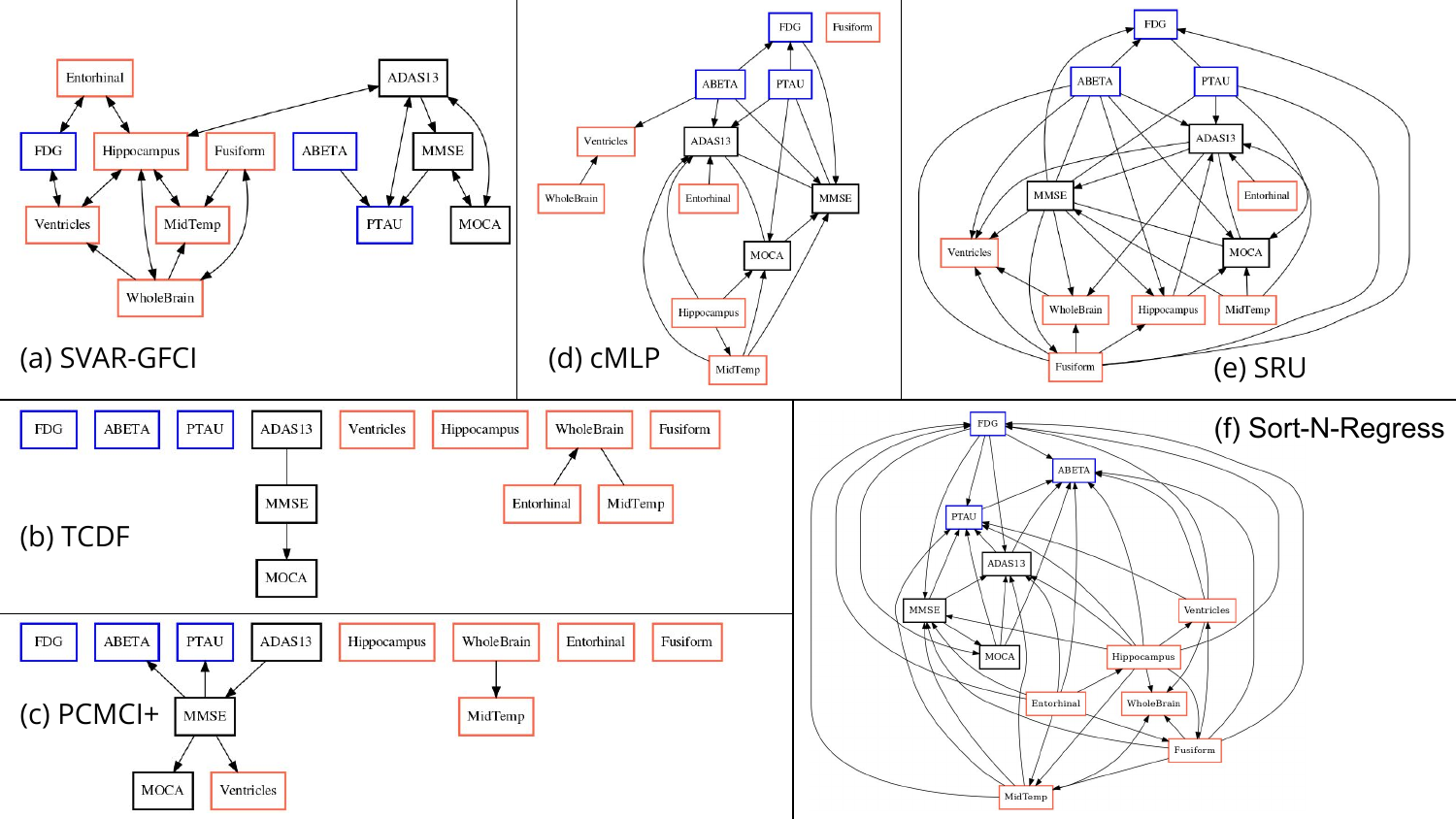}
\caption{Baseline methods' predicted interaction of ADNI biomarkers.
ROI volumes are in red, cognitive tests are in black, and the rest are in blue.
ABETA: amyloid beta, PTAU: phosphorylated tau.
}\label{fig:adni_12var_baselines_app}
\end{figure}

Fig~\ref{fig:adni_12var_baselines_app} shows the outputs of the Sort-N-Regress baseline and the remaining timeseries baselines on the ADNI data.
The output of Sort-N-Regress seems implausible because the nodes FDG, PTAU, and ABETA are children of the MRI nodes.
This contradicts the established literature about Alzheimer's disease in which indicates that the disease seems to originates first from changes in FDG, PTAU, and ABETA (hence these nodes should be roots instead of descendants).
The outputs from PCMCI+ and TCDF are not very informative as they lack edges between the ROIs.
Although somewhat similar \MNAME's output, output from SVAR-GFCI has a lot of bidirectional edges.
The outputs of SRU and cMLP also contain bidirectional edges.
One of the possible reasons for the seemingly worse performance of the baselines is the high missing rate in the ADNI data.

\end{document}